\documentclass{bmvc2k}

\addauthor{Kim Yu-Ji}{ugkim@postech.ac.kr}{1}
\addauthor{Hyunwoo Ha}{hyunwooha@postech.ac.kr}{2}
\addauthor{Kim Youwang}{youwang.kim@postech.ac.kr}{2}
\addauthor{Jaeheung Surh}{jh.surh@bucketplace.net}{3}
\addauthor{Hyowon Ha}{hyowon.ha@bucketplace.net}{3,\footnotemark[2]}
\addauthor{Tae-Hyun Oh}{taehyun@postech.ac.kr}{1,2,4,$\dagger$}

\addinstitution{
 Grad. School of AI\\
 POSTECH, South Korea
}
\addinstitution{
 Dept. of Electrical Engineering\\
 POSTECH, South Korea
}
\addinstitution{
 Bucketplace, Co., Ltd., South Korea
}
\addinstitution{
 Institute for Convergence \\ Research and Education \\in Advanced Technology\\
 Yonsei University, South Korea
}

\runninghead{Yu-Ji et al.}{MeTTA}

\usepackage{graphicx}
\usepackage{caption}

\usepackage[capitalize]{cleveref}
\usepackage{lipsum}
\usepackage{multicol}
\usepackage{makecell}
\usepackage{comment}
\usepackage{float}
\usepackage{listings}
\usepackage{realboxes}
\usepackage{microtype}
\usepackage{amsmath}
\usepackage{bigints}

\usepackage{enumitem}
\usepackage{multirow}
\usepackage{wrapfig}
\usepackage{colortbl}
\usepackage{graphicx}
\usepackage{amssymb}
\usepackage{pifont}
\usepackage{subcaption}
\usepackage{amsmath}
\usepackage{makecell}
\usepackage{xcolor}
\usepackage{listings}
\usepackage{booktabs}

\setlist[itemize]{align=parleft,left=0pt}

\definecolor{azure(colorwheel)}{rgb}{0.0, 0.5, 1.0}
\definecolor{nicegreen}{rgb}{0.0, 0.7, 0.1}
\definecolor{clova}{rgb}{0.24, 0.63, 0.33}
\definecolor{customgray}{rgb}{0.9, 0.9, 0.9}
\definecolor{pink}{cmyk}{0, 0.7808, 0.4429, 0.1412}
\definecolor{amethyst}{rgb}{0.6, 0.4, 0.8}
\definecolor{black}{rgb}{0.0, 0.0, 0.0}
\definecolor{white}{rgb}{1.0, 1.0, 1.0}
\definecolor{red}{rgb}{0.9, 0.0, 0.}
\definecolor{clova}{rgb}{0.24, 0.63, 0.33}
\definecolor{yw}{rgb}{0.01176, 0.5490, 0.5490}
\definecolor{jh}{rgb}{1.0, 0.0, 1.0}
\definecolor{hwha}{rgb}{0.8, 0.5, 0.1}
\definecolor{royalblue}{rgb}{0.25, 0.41, 0.88}

\newcommand{\royalblue}[1]{\textcolor{royalblue}{#1}}
\newcommand{\clovagreen}[1]{\textcolor{clova}{#1}}

\def\eg{\emph{e.g}\bmvaOneDot}
\def\ie{\emph{i.e}\bmvaOneDot}

\newcommand{\ours}{\texttt{MeTTA}}

\newcolumntype{g}{>{\columncolor{customgray}}c}
\newcolumntype{z}{>{\columncolor{customgray}}l}
\newcolumntype{?}[1]{!{\vrule width #1}}

\renewcommand{\paragraph}[1]{\vspace{1mm}\noindent\textbf{#1.}\,\,}

\usepackage{xspace}

\newcommand{\Tref}[1]{Table~\textcolor{blue}{\ref{#1}}}
\newcommand{\Eref}[1]{Eq.~\textcolor{blue}{\ref{#1}}}
\newcommand{\Fref}[1]{Fig.~\textcolor{blue}{\ref{#1}}}
\newcommand{\Sref}[1]{Sec.~\textcolor{blue}{\ref{#1}}}

\newcommand{\bc}{{\mathbf{c}}}

\newcommand{\bk}{{\mathbf{k}}}

\newcommand{\bn}{{\mathbf{n}}}

\newcommand{\bp}{{\mathbf{p}}}

\newcommand{\bx}{{\mathbf{x}}}

\newcommand{\Real}{\mathbb R}

\newcommand{\be}{\begin{eqnarray}}
\newcommand{\ee}{\end{eqnarray}}
\newcommand{\bee}{\begin{eqnarray*}}
\newcommand{\eee}{\end{eqnarray*}}

\newcommand{\matrixb}{\left[ \begin{array}}
\newcommand{\matrixe}{\end{array} \right]}   

\newcommand{\argmin}{\operatornamewithlimits{\arg \min}}

\usepackage{algorithm}
\usepackage{algpseudocode}

\usepackage{hhline}
\usepackage{multirow}
\usepackage{makecell}
\usepackage{graphicx}

\setlist[itemize]{align=parleft,left=0pt,topsep=1mm,itemsep=0mm,parsep=1mm}

\def\eg{\emph{e.g}\bmvaOneDot}

\begin{document}

\title{MeTTA: Single-View to 3D Textured Mesh Reconstruction with Test-Time Adaptation}

\maketitle
\footnotetext[2]{\ denotes corresponding authors.}
\begin{abstract}
Reconstructing 3D from a single view image is 
a long-standing challenge.
One of the popular approaches to tackle this problem is learning-based methods, but dealing with the test cases unfamiliar with training data (Out-of-distribution; OoD) introduces an additional challenge.
To adapt for unseen samples in test time, we propose $\ours$, a test-time adaptation (TTA) exploiting generative prior. 
%
We design joint optimization of 3D geometry, appearance, and pose to handle OoD cases with only a single view image.
However, the alignment between the reference image and the 3D shape via the estimated viewpoint could be erroneous, which leads to ambiguity.
To address this ambiguity, we carefully design learnable virtual cameras and their self-calibration.
In our experiments, we demonstrate that $\ours$ effectively deals with OoD scenarios at failure cases of existing learning-based 3D reconstruction models and enables obtaining a realistic appearance with physically based rendering (PBR) textures.

%
\end{abstract}
\section{Introduction}
Understanding 3D scenes and objects from a single-view image is a long-standing fundamental challenge in computer vision~\cite{marr2010vision}. 
It becomes particularly crucial in robotics for machine perception, extended reality systems for AR/VR, and virtual communication.
They need the ability to comprehend and interact with the real 3D world.
Moreover, representing real 3D scenes requires not only geometric accuracy but also realistic and physically-based properties, essential for creating lifelike and interactive virtual environments~\cite{chen2023fantasia3d, youwang2024paintit}.

\begin{figure}[t]
    \vspace{3mm}
    \centering
    \begin{minipage}{0.48\linewidth}
        \resizebox{1.0\linewidth}{!}{
            \centering
            \includegraphics[width=0.99\linewidth]{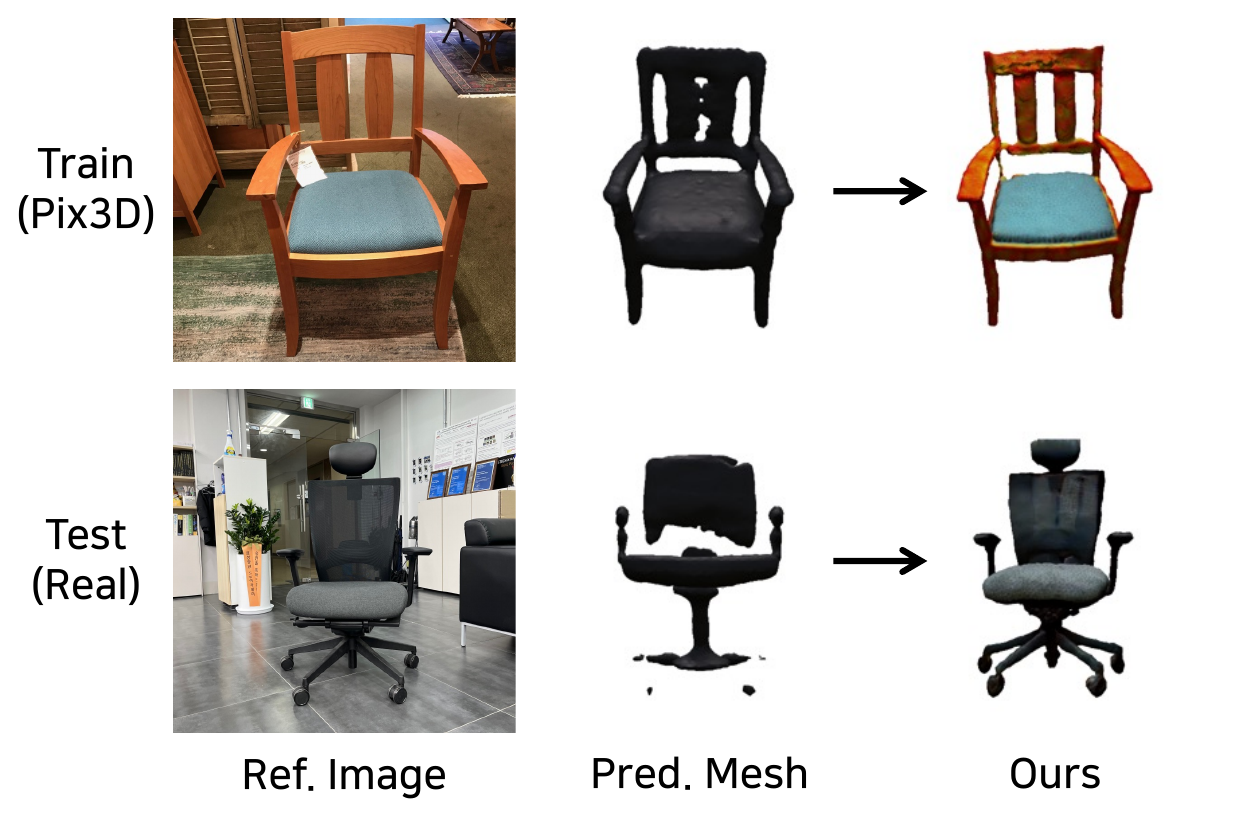}
        }
        \vspace{0.1mm}
        \caption{\textbf{Distribution gap between train and test.}
        ``Train'' refers to a sample on which the Image-to-3D is trained, 
        and ``Test'' is an in-the-wild sample we captured.
        }
        \label{fig:domain_gap}
    \end{minipage}
    \hfill
    \begin{minipage}{0.48\linewidth}
        \resizebox{1.0\linewidth}{!}{
            \centering
            \includegraphics[width=0.99\linewidth]{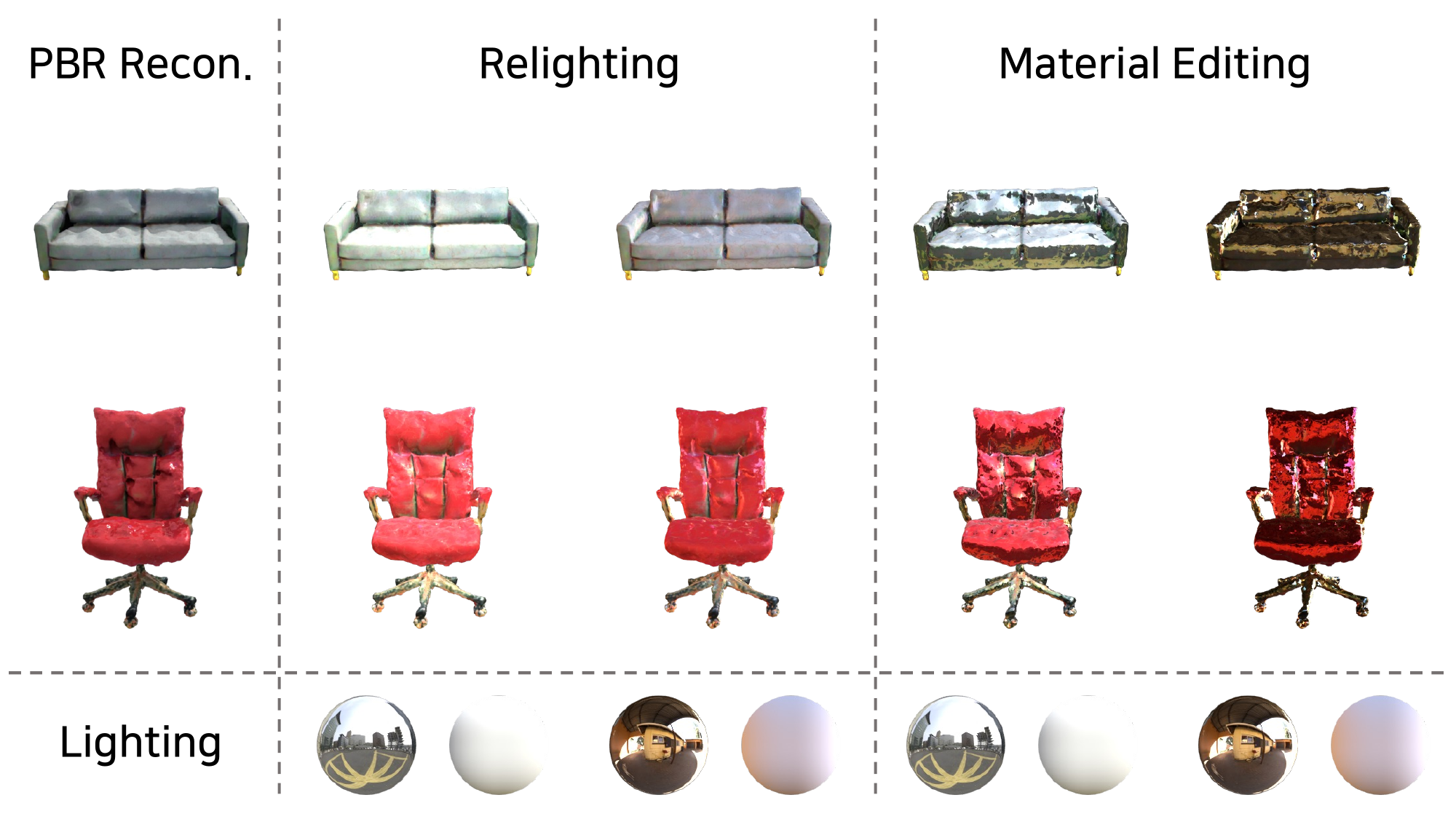}
        }
        \vspace{0.1mm}
        \caption{\textbf{Practical applications in graphics.}
        ``PBR Recon.'' means reconstruction results with PBR textures by ours.
        }
        \label{fig:pbr}
    \end{minipage}
    \vspace{-5mm}
\end{figure}

There have been growing efforts to understand holistic 3d scenes, \eg., layout, object pose, and mesh, from a single-view image~\cite{meshrcnn, total3d, zhang2021holisticim3d, liu2022towardsinstpifu, chen2023singlessr}.
These methods operate effectively by utilizing a learning-based feed-forward approach with reasonable coarse geometry and viewpoint estimation when only given the single-view reference image.
However, the feed-forward methods have the inherent limitation that they cannot perform well on real-world test images away from trained distribution.
Those methods rely on training with \{2D image, 3D shape\}-paired datasets~\cite{sun2018pix3d, fu20213dfront}, which have narrow data distribution compared to the tremendous diversity of real objects.
It is infeasible to construct a large-scale dataset that covers such diversity, considering the difficulty and labor-intensive process of real 3D data acquisition.
Thus, feed-forward methods trained on such a limited dataset can only learn the narrow expressivity of 3D shapes, as shown in ``Pred. Mesh'' of \Fref{fig:domain_gap}.
It hints the vulnerability of such feed-forward models to out-of-distribution (OoD) cases.

To address this challenge, we propose $\ours$, a test-time adaptation (TTA) method for 3D reconstruction by utilizing only a single reference view image.
To compensate for the limited information of single-view, we leverage a pre-trained multi-view generative model~\cite{liu2023zero123} as a prior.
Given a single-view image, we obtain initial mesh and viewpoint predictions from the existing feed-forward model.
We then design joint optimization of the mesh, texture, and camera viewpoint to deal with OoD cases.
However, alignments between the reference image and the 3D mesh from the estimated viewpoint are not exactly matched, which may lead to erroneous results.
To mitigate this, we propose carefully designed learnable virtual cameras with the self-calibrating method to align the 2D pixel information with the 3D shape by updating the initial guess of the viewpoint estimation.

In addition, we parameterize the texture map with physically based rendering (PBR) parameters, including diffuse, specularity, and normal.
This enables us to utilize our results in off-the-shelf graphics tools, \eg, Blender~\cite{blender}; thereby ours can be facilitated to editing for relighting and material control as shown in~\Fref{fig:pbr}.
This is an underexplored feature in previous holistic 3D scene understanding researches~\cite{meshrcnn, total3d, zhang2021holisticim3d, liu2022towardsinstpifu, chen2023singlessr} that predominantly focus on shapes and poses of objects, where we extend to output material property, texture, and mesh complying with input reference image.

Our key contributions are summarized as follows:
\begin{itemize}
    \item We propose $\ours$, which closes the domain gap between training and test time by jointly updating mesh, texture, and viewpoint with the aid of the generative model prior.
    \item We design viewpoint self-calibration and textured mesh reconstruction using only a single view reference image.
    \item We achieve high-fidelity geometry along with a realistic appearance with physically based rendering (PBR) textures, which can be compatible with real graphics engines.
\end{itemize}
\section{Related Work}
Our task is related to the feed-forward reconstruction methods at single-view and the iterative test-time adaptation aided by a generative prior.
We briefly review these lines of work.

\paragraph{Feed-forward reconstruction methods}
This task aims to reconstruct 3D mesh from a single-view image captured in a real-world environment~\cite{zhang2018genre, wu2018shapehd}.
A line of work~\cite{jiajun2017marrnet, meshrcnn, total3d, zhang2021holisticim3d, liu2022towardsinstpifu} have proposed learning-based models that reconstruct image-aligned 3D meshes and poses of objects from a single 2D image.
While they could reconstruct the geometry of objects of given single-view image in an feed-forward manners, they are vulnerable to out-of-distribution (OoD) scenarios beyond the training dataset. 
The out-of-distribution cases for this task are common since the intricacy and the diversity of object shapes in a real-world environment are too complicated to be learned from the limited scale and diversities of existing \{2D image, 3D shape\}-aligned and -paired datasets~\cite{sun2018pix3d, fu20213dfront,collins2022abo,lim2013ikea}.
Moreover, these methods could not represent the texture. 
A recent work~\cite{chen2023singlessr} has explored the reconstruction of 3D mesh and texture from a single image.
However, their feed-forward estimation of shape and texture also could not generalize to real-world cases. 
Also, the model only estimates the RGB color and does not model the physically based rendering (PBR) characteristics, which may limit the realism of the reconstructed texture.

\paragraph{Iterative reconstruction methods using generative priors}
Recent advances in the field of 2D generative models~\cite{stablediffusion, sdxl, sd3, balaji2022ediffi, imagen, dalle, dalle2, dalle3} have shown remarkable capabilities as the prior for 2D inverse problems~\cite{chung2023diffusion,chung2022improving,kawar2022denoising,song2023pseudoinverseguided}.
For our task of single-view 3D textured mesh reconstruction, prior knowledge about 3D object geometry and textures is mandatory to embody a test-time adaptability for OoD cases.
However, directly constructing a 3D object geometry or appearance prior is challenging, considering its unmeasured diversity.

A seminal work, DreamFusion~\cite{poole2022dreamfusion} unlocked the capabilities of a pre-trained text-to-image diffusion model and proposed the Score-Distillation Sampling (SDS), which acts as a 2D generative prior for the 3D generation task~\cite{lin2023magic3d,chen2023fantasia3d,wang2023prolificdreamer,jiang2023avatarcraft}. 
We exploit the idea of using a pre-trained generative model as a prior for 3D tasks. 
Specifically, we propose to use a multi-view diffusion model~\cite{liu2023zero123} as a generative prior to mitigate the test-time distribution shift of the 3D shape, texture and poses.
Additionally, recently proposed feed-forward reconstruction methods
with generative priors~\cite{liu2023one2345, wang2024crm} also cannot model the realistic PBR properties.

\section{Method}
We first provide the overall $\ours$ pipeline in~\Sref{sec:pipeline}.
Following that, we explain how we obtain the coarse object geometry in~\Sref{sec:feedforward} and align the virtual camera to match with the 2D single-view image in~\Sref{sec:3dspace}.
We describe our test-time adaptation (TTA) process for 3D reconstruction in~\Sref{sec:tta} and explain the details of texture representation in~\Sref{sec:pbr}.

\subsection{Overall Pipeline}\label{sec:pipeline}
\begin{figure*}[t]
  \small
  \centering
    \includegraphics[width=0.99\linewidth]{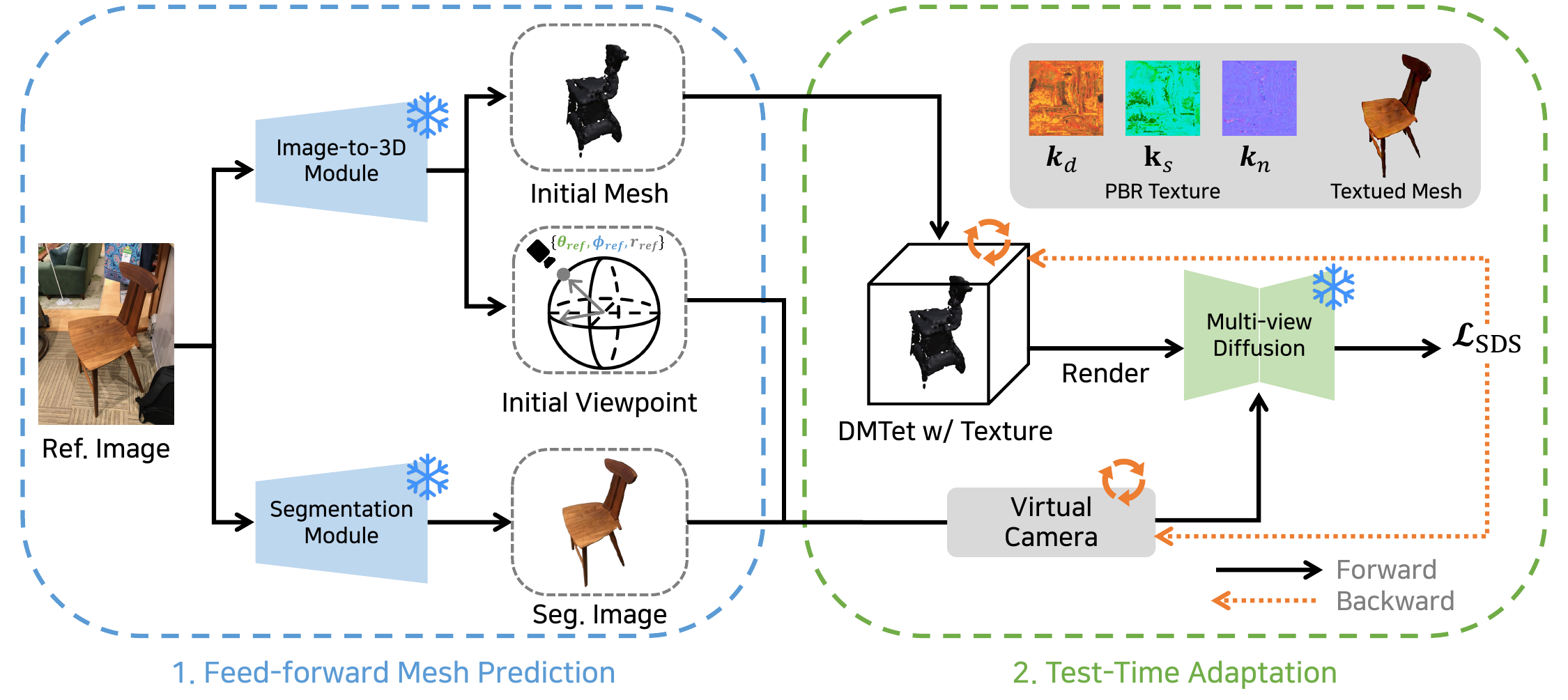}
    \vspace{2mm}
    \caption{\textbf{Overview of $\ours$.} 
    We propose a test-time adaptation pipeline to reconstruct a 3D mesh with PBR texture from a single-view image.
    ``Ref. Image'' refers to the reference input image.
    ``Seg. Image'' refers to the object-segmented image from ``Ref. Image''.
    }
    \vspace{-6mm}
    \label{fig:pipeline}
\end{figure*}

\begin{figure}[t]
    \vspace{3mm}
    \centering
    \begin{minipage}{0.48\linewidth}
        \resizebox{1.0\linewidth}{!}{
            \centering
            \includegraphics[width=0.99\linewidth]{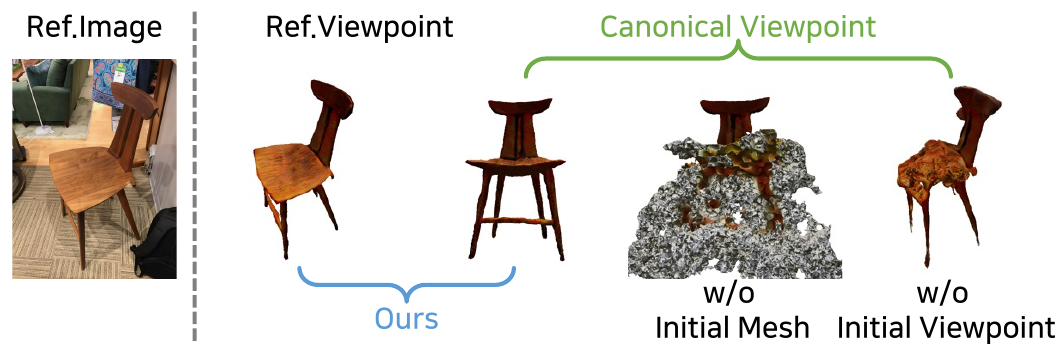}
        }
        \vspace{0.1mm}
        \caption{\textbf{Ablation studies.}
        To validate our pipeline design, we perform ablation studies where the initial mesh or viewpoint prediction is absent.
        In the case of a missing initial mesh, we initialize our 3D space with ellipsoid.
        Canonical viewpoint means that the azimuth and elevation angles are 0$^\circ$.
        }
        \label{fig:abla}
    \end{minipage}
    \hfill
    \begin{minipage}{0.48\linewidth}
        \resizebox{1.0\linewidth}{!}{
            \centering
            \includegraphics[width=0.99\linewidth]{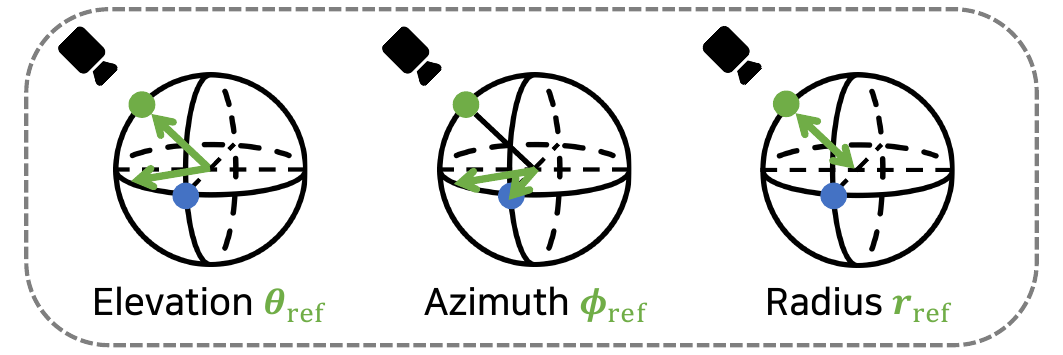}
        }
        \vspace{0.1mm}
        \caption{\textbf{Learnable virtual camera.}
        The reference image is taken with viewpoint (\clovagreen{$\theta_\text{ref}, \phi_\text{ref}, r_\text{ref}$}), which we estimate and optimize.
        \textbf{\clovagreen{Green dot}} means predicted viewpoint given single-view image.
        \textbf{\royalblue{Blue dot}} means canonical viewpoint with both elevation and azimuth angles are 0$^{\circ}$.
        }
        \label{fig:view}
    \end{minipage}
    \vspace{-5mm}
\end{figure}

\vspace{-2mm}
When provided with a single-view reference image during test time, we employ a feed-forward reconstruction method to obtain initial coarse shape and viewpoint predictions in the first stage (\royalblue{blue box}) of~\Fref{fig:pipeline}.
We update coarse geometry to fine-grained shape with realistic textures and viewpoints aligned with a 2D image in the second stage (\clovagreen{green box}) of~\Fref{fig:pipeline}.
We utilize a multi-view diffusion model~\cite{liu2023zero123} to guide the adaptation process through Score-Distillation Sampling (SDS) loss~\cite{poole2022dreamfusion}.
We leverage the segmentation module~\cite{kirillov2023sam, ke2023segmentsamhq, groundedsam} to obtain a white-background object image.
The initial estimated viewpoint has an ambiguity between the 3D object and the reference image.
To mitigate the vagueness, we assume a learnable virtual camera space with its self-calibration which aids in finding well-aligned 2D pixel to 3D space mapping, facilitating seamless adaptation.
We demonstrate the effectiveness of our design, composed of both the initial feed-forward mesh and viewpoint prediction stage and the subsequent test-time adaptation stage, as illustrated in~\Fref{fig:abla}.

\subsection{Feed-forward Initial Prediction}\label{sec:feedforward}
Given a single view input image, we first predict a coarse mesh and its viewpoint by the base Image-to-3D model. 
We can adapt a pre-trained 2D detector (e.g., Faster R-CNN~\cite{girshick2015fast}) into our system, ensuring that it encompasses the specific class we intend to reconstruct.
We then integrate the separate 3D detection and mesh prediction networks that have the 2D detections as input and output SDF representation for mesh and its viewpoint for each object in the input scene, respectively.
We train the 3D networks on the Pix3D~\cite{sun2018pix3d} and SUN RGB-D~\cite{song2015sun} datasets.
We refer to the whole pipeline as the base model~\cite{total3d, zhang2021holisticim3d}.

\subsection{Learnable Virtual Camera}\label{sec:3dspace}
Recall that we obtain predictions for the initial mesh and camera viewpoint (e.g., radius, elevation and azimuth angles) using the feed-forward model.
At test time, the camera parameters of camera focal length and pose parameters are unknown, leading to the ambiguity between 2D pixel information and 3D shape mapping.
To address this ambiguity, we define a learnable virtual camera, where we set pre-defined camera intrinsics and adapt the extrinsic pose of the virtual camera.
We need refinement to align the mapping because the viewpoint estimation from the previous step is just an initial guess and may be erroneous.

Getting aligned 3D mesh to 2D image observation is essential to utilize multi-view diffusion priors.
In the pre-optimization stage, we set the initial viewpoint from these predictions and first update the radius of our virtual camera by optimizing the initial mesh rendering to be aligned with the reference image with mask loss.
In the main optimization stage, we propose to self-calibrate the virtual camera pose by simultaneously optimizing our 3D mesh with PBR texture to achieve a more accurate alignment between the 2D image and the 3D space.
We estimate and update the reference viewpoint ($\theta_{ref}, \phi_{ref}, r_{ref}$) to align between 2D reference image and the 3D shape, as shown in~\Fref{fig:view}.
This approach refines the mapping between a 2D image and 3D space and obtains consistent 3D results, which is vital for holistic scene reconstruction.
Based on the reference viewpoint, we sample the relative viewpoint ($\Delta\theta, \Delta\phi, \Delta r$) as a condition to the multi-view diffusion model~\cite{liu2023zero123}.

\subsection{Test-Time Adaptation for 3D Reconstruction.}\label{sec:tta}
We employ DMTet~\cite{shen2021deepdmtet} as our 3D representation, which is characterized by two essential features; a deformable tetrahedral grid used to represent 3D shapes and a differentiable marching tetrahedral (MT) layer designed to extract explicit triangular meshes.
DMTet has $V_T$ vertices in the tetrahedral grid $T$, which can be expressed as $(V_T, T)$.

\paragraph{DMTet initialization from coarse geometry}
To model the geometry and texture of a 3D object, for each vertex $v_i \in V_T$, we learn the signed distance function (SDF) $s(v_i)$, vertex deformation offset $\Delta v_i$ and per-vertex physically based rendering (PBR) material properties $\bk_\text{PBR}$, with hash-grid positional encoding~\citep{muller2022instant} function $\tau$ as follows:
\begin{equation}\label{eqn:dmtet}
    [s(v_i), \Delta v_i, \bk_\text{PBR}] = \Theta (\tau(v_i) ;\theta),
\end{equation}
where MLP network $\Theta$ has the parameters $\theta$. 
Before optimizing the target object from the reference image, we initialize DMTet with the initial shape obtained from the base model.
From this initial mesh, we randomly sample a set of points $\{p_i \in \Real^3\}$ where $p_i$ represents a point in $P$ which is the mesh vertices. 
We initialize the DMTet grid and its neural parameters to fit the initial mesh prediction by solving a SDF optimization problem as follows:
\begin{equation}
   \theta^{*} = \argmin_{\theta}\sum_{p_i \in P} \Vert s(\tau(p_i);\theta) - \text{SDF}(p_i) \Vert_2^2.
\end{equation}
Using the pre-optimized network $\Theta$ and a differentiable renderer $R$, \eg, Nvdiffrast~\cite{nvidiffrast}, we obtain the RGB rendering image $\mathbf{x}$ as $\mathbf{x} =  R(\theta, c)$,
where $c$ represents the sampled camera viewpoint.
We randomly sample camera viewpoints within the range of [-45$^\circ$, 45$^\circ$] for the elevation angle and [0$^\circ$, 360$^\circ$] for the azimuth angle.

\paragraph{Jointly optimizing shape, texture \& camera}
Given the initialized DMTet and its corresponding MLP $\Theta$, we proceed to adapt the shape, texture and the virtual camera pose jointly.
To update $\Theta$ parameterized by $\theta$, we utilize Score-Distillation Sampling (SDS) loss, which calculates per-pixel gradients by computing the difference between predicted noise and added noise as follows:
\begin{equation}
    \nabla_\theta \mathcal{L}_\text{SDS}(\psi, \mathbf{x}) = \mathbb{E} \biggl[ w(t)(\mathbf{\epsilon}_\psi(\mathbf{z}_t; \mathbf{y}, t) - \mathbf{\epsilon} )\frac{\partial \mathbf{z}}{\partial \mathbf{x}}\frac{\partial \mathbf{x}}{\partial \theta} \biggr],
\end{equation}
where $\psi$ parameterizes multi-view aware image diffusion model, $\mathbf{x}$ represents the RGB rendering output, $w(t)$ signifies a weight function for different noise levels, $\mathbf{z}_t$ denotes the latent encoding of $\mathbf{x}$ with the addition of noise $\mathbf{\epsilon}$, and $\mathbf{\epsilon}_{\psi}$ is the predicted noise with reference image $\mathbf{y}$ and noise level $t$.

We leverage several additional loss terms to aid in the optimization.
To promote the photometric consistency between the reference image and rendered textures of the 3D reconstruction, we introduce the photometric loss $\mathcal{L}_\text{photo} = \Vert I_\text{ref} - \mathbf{x}_\text{ref}  \Vert_1$ between the reference image $I_\text{ref}$ and the rendering from the reference viewpoint $\mathbf{x}_\text{ref}$.
Similar to the photometric loss, we also leverage the mask loss $\mathcal{L}_\text{mask} = \Vert M(I_\text{ref}) - M(\mathbf{x}_\text{ref})  \Vert_1$, which $M$ is the masking function used for binary separation between the object and the background.
It compares the mask of the reference image with the mask of the rendering to promote shape consistency.

To impose regularization on the mesh surface, parameterized by SDF representations, we employ SDF regularization methods akin to those proposed by \citet{liao2018deepmc} and \citep{li2023neuralangelo}.
Utilizing the binary cross entropy ($BCE$), the sigmoid function $\sigma$, and the sign function, we can express the SDF regularizer $\mathcal{L}_\text{reg} = \sum_{(i, j) \in \mathbb{S}} \Bigl(BCE(\sigma(s_i), \text{sign}(s_j)) + BCE(\sigma(s_j), \text{sign}(s_i) \Bigr)$,
where $s_i$ is the SDF value at the vertex $v_i$ and $\mathbb{S}$ is set of unique edges.
To further encourage the smoothness of the reconstructed surface, we regularize the mean curvature of SDF, which can be computed from discrete mesh Laplacian.
The Laplacian loss is defined as $\mathcal{L}_\text{lap} = \frac{1}{N} \sum_{i=1}^N | \nabla^2 s_i |$.
The overall loss can be defined as the combination of $\mathcal{L}_\text{SDS}, \mathcal{L}_\text{photo}, \mathcal{L}_\text{mask}, \mathcal{L}_\text{reg}$ and $\mathcal{L}_\text{lap}$.
We backpropagate the losses to jointly update the 3D shape, PBR texture, and poses of the learnable virtual camera.

\subsection{Neural PBR Texture Optimization}\label{sec:pbr}
As aforementioned in \Eref{eqn:dmtet},
we employ DMTet in conjunction with a physically based rendering (PBR) material model~\citep{mcauley2012practicalpbr}, similar to~\citep{nvdiffrec}. 
This choice allows us to incorporate spatially-varying Bidirectional Reflectance Distribution Function (BRDF) modeling for textures, yielding a more realistic appearance.
The PBR material properties, $\bk_\text{PBR}$ is composed of three key components: diffuse lobe parameters $\bk_d \in \mathbb{R}^3$, the roughness and metalness term $\bk_{rm} \in \mathbb{R}^2$, and the normal variation term $\bk_n \in \mathbb{R}^3$.
The specular highlight color, denoted as $\bk_s \in \mathbb{R}^3$, can be determined with the renowned Cook-Torrance microfacet BRDF model~\cite{cook1982rendering}. 
Given diffuse value $\bk_d$ and the metalness factor $m$, we compute $\bk_s$ as: $\bk_s = (1 - m) \cdot 0.04 + m \cdot \bk_d$.
It enables us to achieve photorealistic surface rendering and enhances the potential of diffusion models for improved realism. More details are in the supplementary material.
\section{Experiments}
In this section, we first explain the experimental setup in~\Sref{sec:setup}.
Following that, we show the verification of our system design choices (\eg., virtual camera and test-time adaptation) in~\Sref{sec:verify}.
We demonstrate our high-fidelity textured mesh reconstruction results in respect of quality and quantity in~\Sref{sec:qual} and~\Sref{sec:quan}, respectively.

\subsection{Experimental Setup}\label{sec:setup}
To evaluate the cross-domain robustness of $\ours$'s 3D reconstruction performance, we conduct experiments on the 3D-Front dataset~\cite{fu20213dfront}, which has not been used in previous single-view to 3D reconstruction methods~\cite{total3d, zhang2021holisticim3d}, and we select fifteen samples for evaluation.
To demonstrate that our pipeline is working in real-world, out-of-domain scenarios, we manually acquire images from the real scene and the web.
For in-domain evaluations, we extract a subset from the Pix3D dataset~\cite{sun2018pix3d}.
Due to time complexity considerations at the optimization, we had to limit the number of dataset selections to a few dozen.

\subsection{Verification of System}\label{sec:verify}
In this section, we show the experiments to verify the effectiveness of our system design choices, especially for the learnable virtual camera and the test-time adaptation stage.

\begin{figure}
    \centering
    \begin{minipage}{0.48\linewidth}
        \resizebox{1.0\linewidth}{!}{
            \centering
            \includegraphics[width=0.99\linewidth]{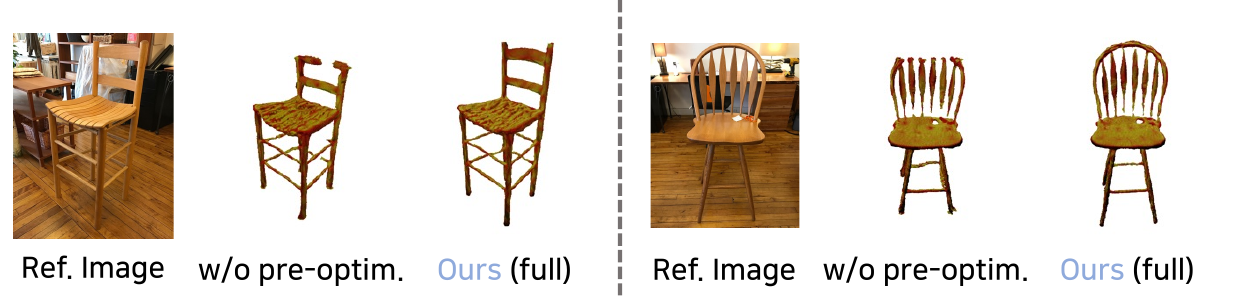}
        }
        \vspace{0.1mm}
        \caption{\textbf{Necessity of pre-optimization for radius.}
        The ``w/o pre-optim.'' cases exhibit geometry cut-off for exceeding the camera space boundary and degradation of details.
        }
        \label{fig:radius_optim}
    \end{minipage}
    \hfill
    \begin{minipage}{0.50\linewidth}
        \resizebox{1.0\linewidth}{!}{
            \begin{tabular}{c cc}
            \toprule
                Metric &  Ours (w/o self-calibration) & Ours (full) \\
            \midrule
                Chamfer Distance $\downarrow$ & 0.0593 & \textbf{0.0580} \\
                F-Score (\%) $\uparrow$ & 50.35 & \textbf{51.15} \\
            \bottomrule
            %
            %
            \end{tabular}
        }
        \vspace{2mm}
        \captionof{table}{\textbf{Effectiveness of self-calibration for angles.}
        Ours (full) shows better consistency, depicting the self-calibration effectiveness.
        We average over all fifteen samples.
        }
        \label{tab:self_calibrate}
    \end{minipage}
    \label{fig:vefiry}
    \vspace{-5mm}
\end{figure}

\paragraph{Effectiveness of learnable virtual camera}
We show the ablation studies of camera pre-optimization and self-calibration.
The pre-optimization stage is crucial to find the proper radius scale for detailed structures, as shown in~\Fref{fig:radius_optim}.
We also present an ablation study of the camera self-calibration in~\Tref{tab:self_calibrate}.
We add angle perturbations of [-15, -10, -5, 5, 10, 15] degrees to initial viewpoint estimations.
Then, we measure the average scores of the results 
with respect to
the 3D mesh obtained with no perturbation.
The self-calibration stage is essential to refine the mapping between a 2D image and 3D space and obtain physically accurate and consistent 3D results, which is vital for total scene reconstruction.

\begin{wrapfigure}{r}{0.5\linewidth}
    \small
    \vspace{-6mm}
    \centering
        \includegraphics[width=0.99\linewidth]{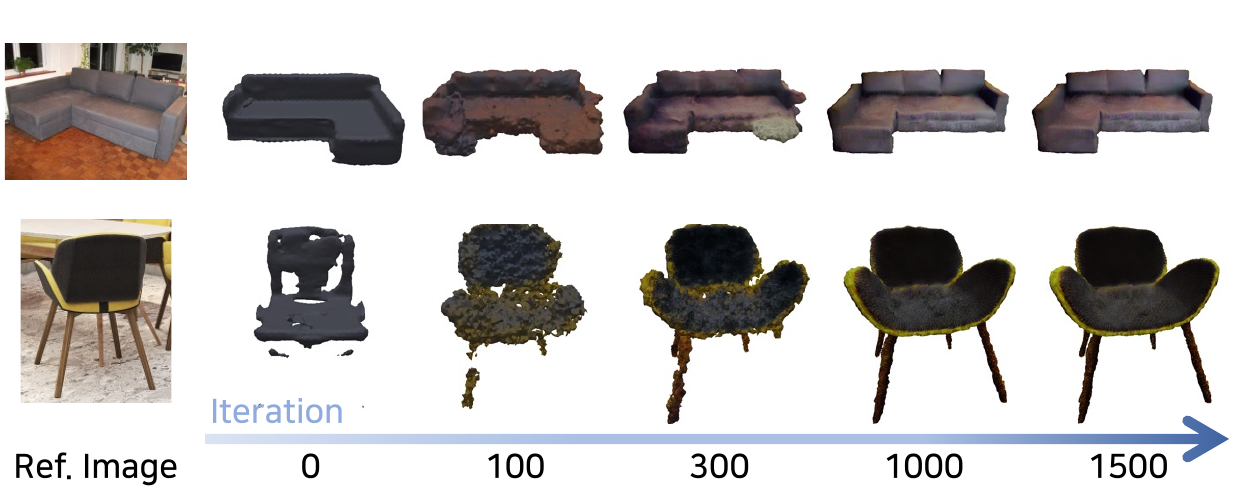}
        \vspace{1mm}
        \caption{
        \textbf{Intermediate results.}
        Our method robustly refines meshes and textures iteratively, even with poor initialization.
        }
    \label{fig:intermediate}
\end{wrapfigure}

\paragraph{Effectiveness of test-time adaptation}
We show the intermediate iteration results during the second stage to present the 
necessity of the test-time adaptation (TTA)
in~\Fref{fig:intermediate}.
While bad initials occur quite often in the Image-to-3D module due to an out-of-distribution gap between training and test, the intermediate results clearly show the strength and necessity of our second TTA stage.

\begin{figure}[t]
  \centering
    \includegraphics[width=0.95\linewidth]{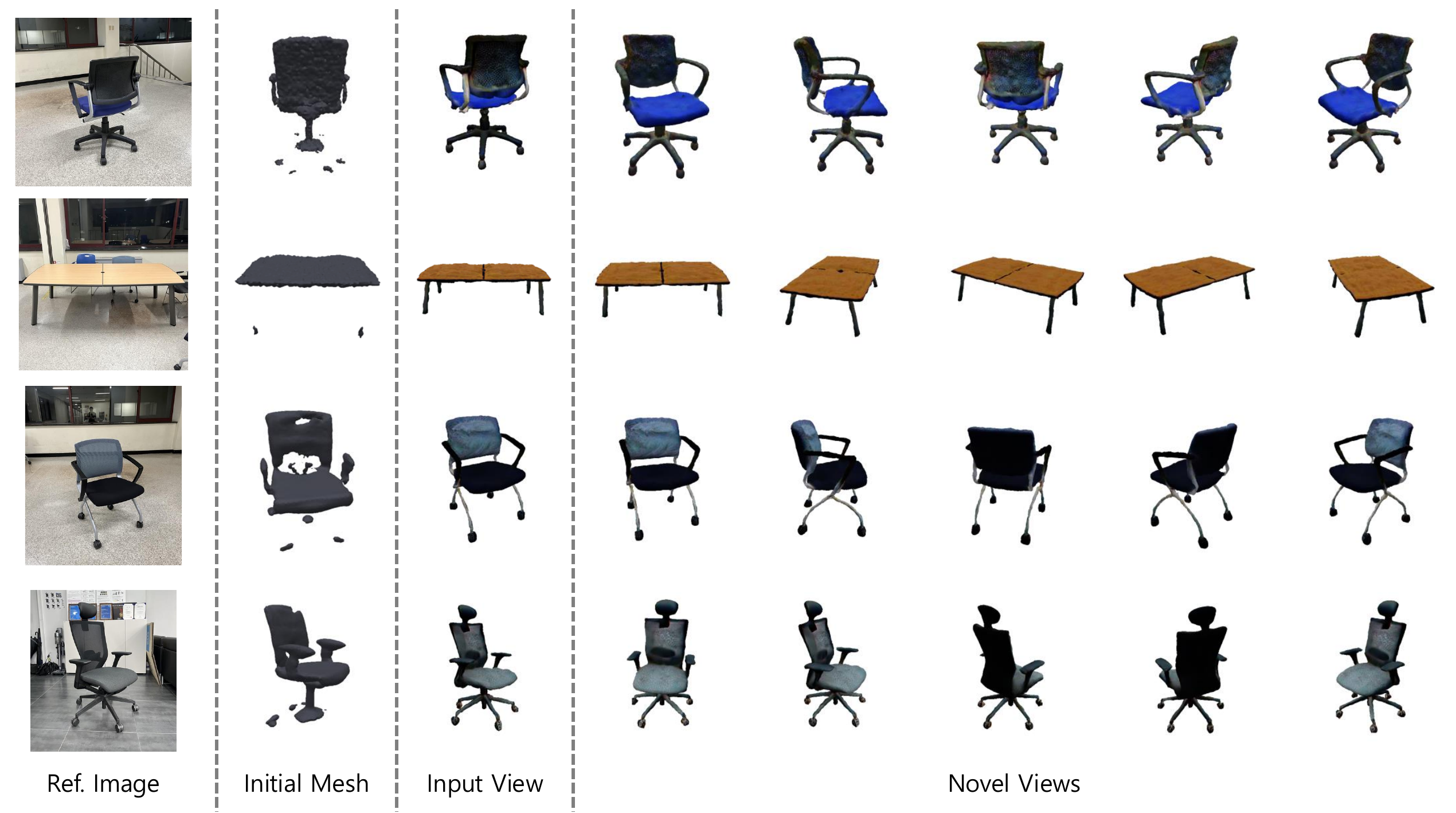}
    \vspace{2mm}
    \caption{\textbf{Unseen real-world experiments about manually acquired data.} 
    We showcase the effectiveness of our test-time adaptation for real scenarios.
    }
  \label{fig:in_the_wild}
\end{figure}
\begin{figure}[h!]
  \centering
    \includegraphics[width=0.93\linewidth]{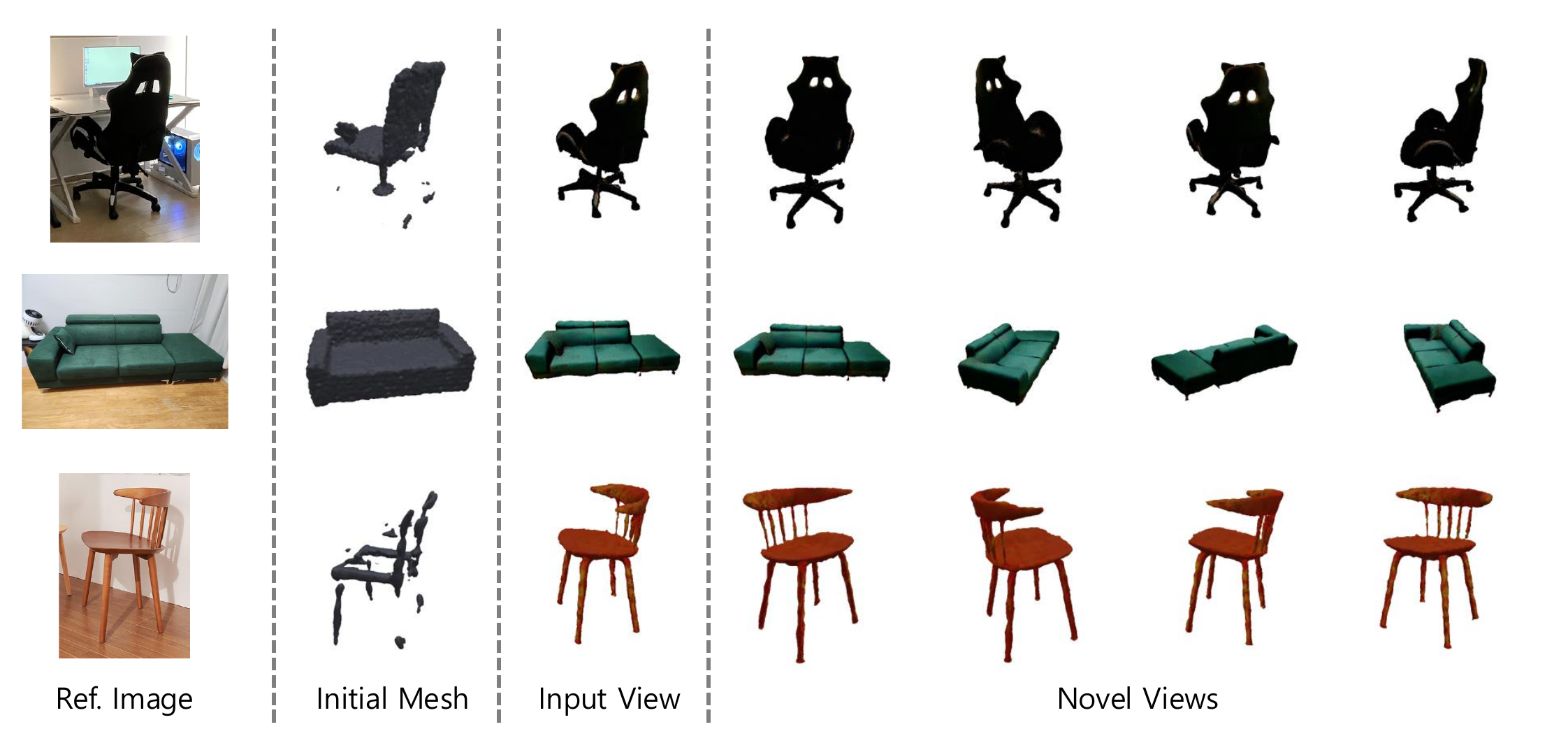}
    \vspace{2mm}
    \caption{\textbf{Unseen real-world experiments about in-the-wild web images.} 
    We showcase the effectiveness of our test-time adaptation for real scenarios.
    }
  \label{fig:ohouse}
\end{figure}
\begin{figure}[t]
  \centering
    \includegraphics[width=0.95\linewidth]{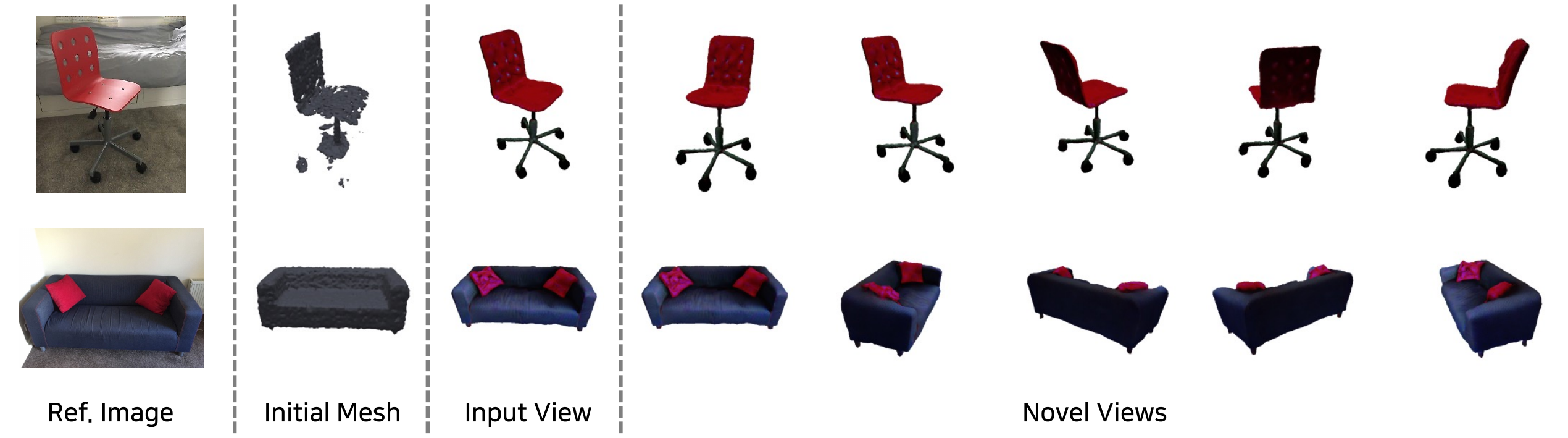}
    \vspace{2mm}
    \caption{\textbf{In-domain experiments.} 
    We showcase the effectiveness of our test-time adaptation of in-domain datasets in which the Image-to-3D module is trained.
    }
    \vspace{-5mm}
  \label{fig:in_domain}
\end{figure}

\subsection{Qualitative Analysis}\label{sec:qual}
We evaluate and compare the 3D mesh reconstruction quality of $\ours$ with the competing methods.
For more qualitative results, please refer to the supplementary material.

\paragraph{Textured 3D mesh reconstruction}
We assess the quality of reconstructed 3D textured meshes in terms of geometric and appearance attributes.
In~\Fref{fig:in_the_wild}, our results show notable achievement, where we can reconstruct a realistically textured novel-view 3D mesh only from a partial observation of the 3D object in the previously unseen scenarios.
In~\Fref{fig:ohouse}, we conduct another real-world experiment about web images and show fine-grained detailed 3D textured mesh reconstruction results.
In~\Fref{fig:in_domain}, feed-forward methods~\cite{total3d, zhang2021holisticim3d} predict the coarse geometry corresponding to the reference image to some extent.
However, for detailed geometry and realistic texture, it is essential to apply our test-time adaptation process, even for the in-domain settings.

\begin{figure}[t]
    \begin{minipage}{0.36\linewidth}
        \centering
        \small
        \resizebox{0.99\linewidth}{!}{
          \small
          \centering
            \includegraphics[width=0.99\linewidth]{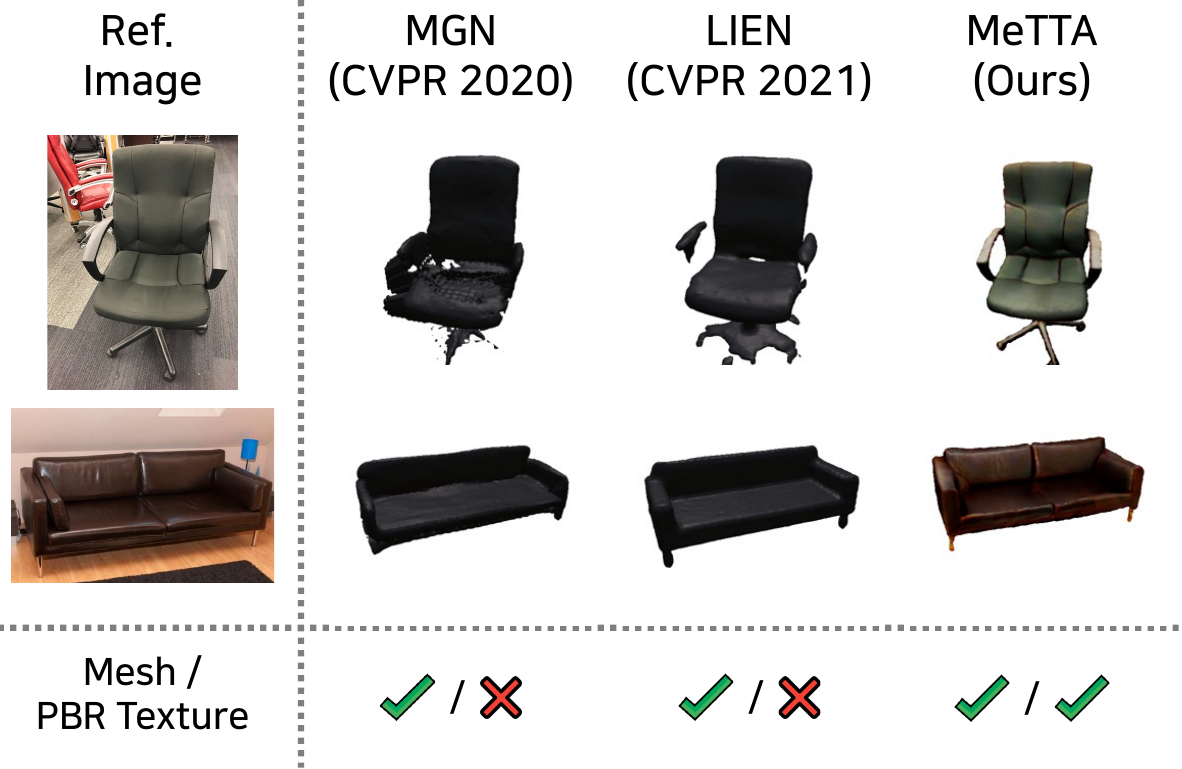} 
        }\vspace{3mm}
        \caption{\textbf{Comparison with feed-forward methods.} 
        }
        \label{fig:count_tta}
    \end{minipage}
    \hfill
    \begin{minipage}{0.62\linewidth}
      \centering
      \resizebox{0.99\linewidth}{!}{
        \centering
        \includegraphics[width=0.99\linewidth]{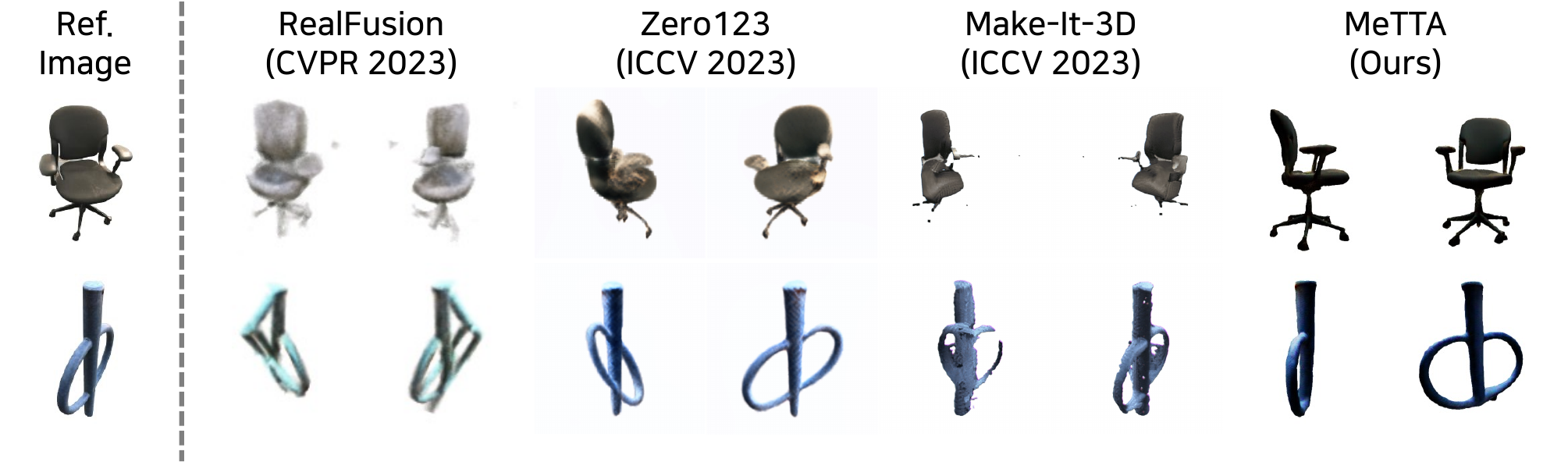}
        }
      \vspace{1mm}
      \caption{\textbf{Comparison with iterative methods using generative priors.} 
      Ours show photo-realistic texture details with physically accurate geometry.
      }
      \label{fig:count_diff}
    \end{minipage}%
\vspace{-5mm}
\end{figure}

\paragraph{Comparison with feed-forward methods}
We compare ours to previous feed-forward reconstruction methods~\cite{total3d, zhang2021holisticim3d} for visual quality.
Thanks to the test-time adaptation with multi-view generative prior, we can get accurate 3D shapes with realistic PBR textures, as shown in~\Fref{fig:count_tta}.

\paragraph{Comparison with iterative methods using generative priors}
We compare our single image to 3D reconstruction results to existing generative priors methods~\cite{melas2023realfusion, liu2023zero123,tang2023makeit3d}.
Because previous methods do not deal with viewpoint information as our learnable virtual cameras, their 3D reconstruction results are not aligned with the reference image and show distorted results, as shown in~\Fref{fig:count_diff}.

\subsection{Quantitative Analysis}\label{sec:quan}

We also conduct quantitative comparisons to assess the quality of textured mesh reconstruction and the effectiveness of geometric properties.

\paragraph{Comparison with feed-forward methods}   
We compare ours to feed-forward reconstruction methods~\cite{total3d, zhang2021holisticim3d} which are also the base models to evaluate whether they have a valid and accurate 3D structure.
We evaluate the Chamfer Distance of sampled points between the ground-truth mesh and output mesh of each method.
In \Tref{tab:quan_geo}, $\ours$ outperforms geometry reconstruction than competing methods.
Note that our optimization process does not access the ground-truth 3D information, \eg, point clouds, voxels, and meshes, while previous methods are trained to minimize Chamfer Distance with ground-truth 3D shapes as direct supervision.
Note that $\ours$ also reconstruct fine-grained geometries with utilizing only 2D reference image, compared to others which are trained with 3D shape dataset~\cite{sun2018pix3d}.

\paragraph{Comparison with iterative methods using generative priors}  
We compare the texture reconstruction quality of $\ours$
with the competing methods:
RealFusion~\citep{melas2023realfusion}, Zero-1-to-3~\cite{liu2023zero123} and Make-It-3D~\citep{tang2023makeit3d}.
In~\Tref{tab:quan_tex}, 
we measure the similarity between the reference image and the rendered image at the reference view and novel views, respectively.
We use three metrics: PSNR, LPIPS~\citep{zhang2018unreasonable}, and CLIP score~\citep{clip}.
The CLIP score evaluates the semantic similarity.
To see the appearance consistency between novel views, we also report the minimum value of the CLIP score.
\ours~mostly outperforms the competing methods in both reference view and novel view rendering qualities.
The results highlight the 
$\ours$'s capability of preserving the semantics of 3D objects, even for the occluded novel views, while achieving high-fidelity 3D reconstruction.
\vspace{-4mm}

\begin{figure}[t]
    \vspace{3mm}
    \centering
    \begin{minipage}{0.43\linewidth}
        \resizebox{1.0\linewidth}{!}{
            \begin{tabular}{c cccc}
            \toprule
                Metric &  MGN~\cite{total3d} & LIEN~\cite{zhang2021holisticim3d} & \ours~(Ours) \\
            \midrule
                Chamfer Distance $\downarrow$   & 0.1089 & 0.0975 & \textbf{0.0943} \\
            \bottomrule
            \end{tabular}
        }
        \vspace{2mm}
        \captionof{table}{\textbf{Cross-domain evaluation of the single-view to mesh methods.} 
        We evaluate on unseen test dataset~\cite{fu20213dfront}.
        }
        \label{tab:quan_geo}
    \end{minipage}
    \hfill
    \begin{minipage}{0.55\linewidth}
        \resizebox{1.0\linewidth}{!}{
            \begin{tabular}{l  ccc  cc}
            \toprule
                \multirow{2}[2]{*}{Method} & \multicolumn{3}{c}{Reference View} & \multicolumn{2}{c}{Novel Views} \\ \cmidrule(lr){2-4} \cmidrule(lr){5-6}
                & LPIPS $\downarrow$ & PSNR [dB] $\uparrow$ & CLIP Score $\uparrow$ &  CLIP Score $\uparrow$ & min. CLIP Score $\uparrow$\\ 
            \midrule
                RealFusion~\cite{melas2023realfusion}   & 0.1809 & 21.56 & 0.8494 & 0.7538 & 0.7030 \\
                Zero-1-to-3~\cite{liu2023zero123}   & 0.1079 & \textbf{23.53} & 0.9170 & 0.7661 & 0.6670 \\
                Make-It-3D~\cite{tang2023makeit3d}   & 0.0867 & 22.45 & 0.9386 & 0.8937 & 0.8046 \\
                $\ours$ (ours)       & \textbf{0.0777} &  22.89 & \textbf{0.9465} & \textbf{0.8942} & \textbf{0.8286} \\
            \bottomrule
            \end{tabular}
        }
        \vspace{2mm}
        \captionof{table}{\textbf{Comparisons of texture reconstruction and perceptual quality.}
        }
        \label{tab:quan_tex}
    \end{minipage}
    \vspace{-6mm}
\end{figure}

\section{Discussion, Limitation, and Conclusion}
In this work, we present $\ours$, a monocular 3D textured mesh reconstruction with generative test-time adaptation.
Our approach addresses several challenges in reconstructing a 3D textured mesh from a single image. 
First, we highlight the limitations of single-view to 3D mesh prediction methods based on feed-forward manners, which often struggle to ensure high-quality mesh estimation results due to limited 3D shape representation learned from the existing closed training set. 
Second, we emphasize the necessity of self-calibrating the learnable virtual camera to connect
different coordinate spaces between Image-to-3D shape models and the multi-view image generative prior model.
Tackling the challenges enables us to achieve quality geometry and photo-realistic texture appearance, complying with input.
Finally, We discuss our limitations and conclude with future directions.

\paragraph{Optimization-based system}
Ours is much faster than fair competitors, optimization-based approaches~\cite{melas2023realfusion, tang2023makeit3d}.
Specifically, our test-time adaptation stage takes 30 minutes per object, compared to 193 minutes of RealFusion~\cite{melas2023realfusion} and 91 minutes of Make-It-3D~\cite{tang2023makeit3d}.
However, we acknowledge that there is still work to achieve practicality, especially in real-time.

\begin{wrapfigure}{r}{0.5\linewidth}
    \small
    \vspace{-6.5mm}
    \centering
        \includegraphics[width=0.99\linewidth]{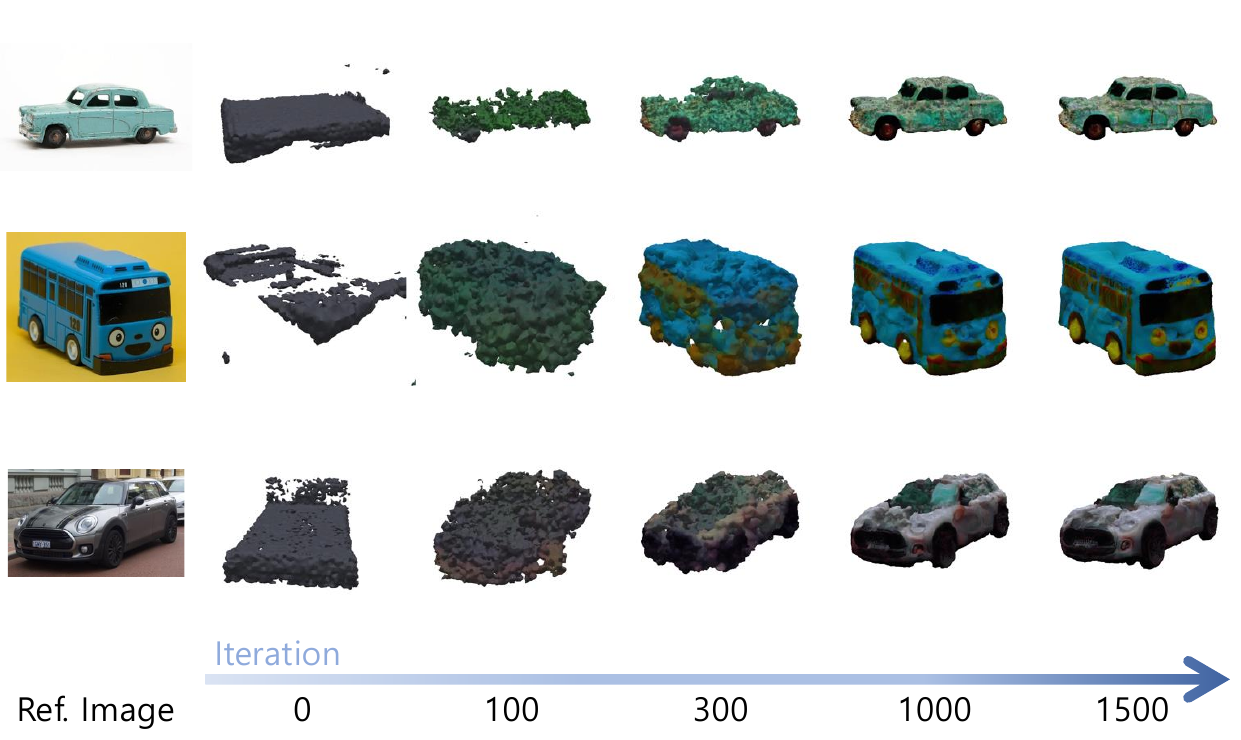}
        \vspace{0.3mm}
        \caption{
        \textbf{Possibility of category extension.}
        Because the Image-to-3D module is trained with 9 indoor object classes~\cite{sun2018pix3d}, it predicts the image as a “bed” rather than a “car”.
        }
    \label{fig:general_class}
    \vspace{-6mm}
\end{wrapfigure}
\paragraph{Category generalization}
Our definition of~``cross-domain''~implies training and testing on different datasets within the same intra-category, \eg, furniture to furniture.
Trained on a small-scale 3D dataset~\cite{sun2018pix3d}, our Image-to-3D module's prediction is category-specific.
Despite this, testing in an inter-category scenario in~\Fref{fig:general_class} shows our method is reasonably effective, albeit not designed for such cases.

\paragraph{Future direction}
%
Our two-stage optimization method could be integrated into an end-to-end approach for improved speed and performance.
Enhancing the Image-to-3D stage with more data may improve category generalization. 
We aim to investigate this in future work.


\paragraph{Acknowledgment}
This project was supported by Bucketplace and also supported by the Institute of Information \& communications Technology Planning \& Evaluation (IITP) grant funded by the Korea government(MSIT) (No.RS-2022-II220290, Visual Intelligence for Space-Time Understanding and Generation based on Multi-layered Visual Common Sense; No.RS-2022-II220124, Development of Artificial Intelligence Technology for Self-Improving Competency-Aware Learning Capabilities; and No.RS-2019-II191906, Artificial Intelligence Graduate School Program(POSTECH)).

\bibliography{egbib}

\clearpage

\runninghead{Yu-Ji et al.}{MeTTA}

\renewcommand{\thefigure}{S\arabic{figure}}
\renewcommand{\thetable}{S\arabic{table}}

\crefname{section}{Sec.}{Secs.}
\Crefname{section}{Section}{Sections}
\Crefname{table}{Table}{Tables}
\crefname{table}{Tab.}{Tabs.}


\appendix
\section*{Supplementary Material}
This supplementary material presents technical details, analyses, and experiments not included in the main paper due to the space limit.

\section{Technical Details}
This section provides detailed information on the implementation details of the overall pipeline and physically-based rendering (PBR) modeling in the main paper.

\subsection{Implementation Details}

\paragraph{Experimental details}
We use AdamW optimizer with gradient clipping and the respective learning rates of 1$\times$10$^{-3}$ for geometry and 1$\times$10$^{-3}$ for texture and optimize them simultaneously.
We randomly sample 8 camera viewpoints for each iteration for rendering the novel views.
We conduct training with one NVIDIA A6000 GPU for about 30 minutes.
We leverage Open3D~\citep{zhou2018open3d} to deal with SDF and point cloud representations.

\paragraph{Chamfer Distance}
We measure Chamfer Distance to assess the quality of the mesh reconstruction.
Point clouds are normalized in scale and aligned to the ground-truth point clouds by the iterative closest point (ICP) algorithm.
10K points are sampled for evaluating each mesh.

\paragraph{Image-to-3D module}
We require a learning-based feed-forward mesh prediction stage employing the Image-to-3D module to obtain a preliminary coarse mesh and initial viewpoint of the input image.
The Image-to-3D module encompasses various techniques capable of predicting a coarse mesh and an approximate viewpoint for the input image, \eg,~\cite{total3d, zhang2021holisticim3d}.

\paragraph{Segmentation module}
\ours~harnesses the multi-view diffusion model~\cite{liu2023zero123} fine-tuned on large-scale synthetic datasets~\cite{deitke2023objaverse, deitke2023objaversexl}, specifically designed for object rendering against a white background. 
Achieving precise object segmentation is pivotal for effectively leveraging the multi-view diffusion model, biased towards images with segmented white backgrounds.
To automate the process of obtaining high-quality segmentation results, we make use of the latest segmentation models~\cite{kirillov2023sam, ke2023segmentsamhq}.
While these models offer substantial automation, they still require some level of user-interactive querying.
In response, we have integrated a grounding method~\cite{liu2023groundingdino} to obtain appropriate object detection as a query.
Based on the detection results as a user-given query, we subsequently employ a user-interactive segmentation method to finalize the fine-grained segmentation results.

\subsection{Texture Modeling}
As explained in Section 3.5. of the main paper, we adopt physically-based rendering (PBR) material modeling~\cite{mcauley2012practicalpbr} to optimize neural texture optimization.
By employing PBR material modeling, we can achieve a realistic appearance for the reconstructed object and easily integrate it with various graphics engines (\eg, Blender~\cite{blender}) for practical applications.
The PBR material properties, denoted as $\bk_\text{PBR}$, consist of three fundamental elements: diffuse lobe parameters $\bk_d \in \mathbb{R}^3$, the roughness and metalness term $\bk_{rm} \in \mathbb{R}^2$, and the normal variation term $\bk_n \in \mathbb{R}^3$.
$\bk_{rm}$ consists of the roughness $r$ and metalness term $m$.
The first term, $r$, is a parameter of GGX~\cite{walter2007microfacet} normal distribution function and affects how the material's surface reflects light.
The second term, $m$, is used with diffuse value $\bk_d$ for computing the specular term $\bk_s = (1 - m) \cdot 0.04 + m \cdot \bk_d$. 
We employ a tangent space normal map, denoted as $\bk_n$, to capture intricate high-frequency lighting details on the surface.
With a given scene environment light~\cite{poliigon}, we can compute a basic rendering equation as a basic image-based lighting model denoted by:
\begin{equation}
    L_{\theta}(\bp, \bc) = \int_{\Omega}L_i(\bp, \bc_i)f_{\theta}(\bp, \bc_i, \bc)(\bc_i \cdot \bn_{\bp})d\bc_i,\label{eq:light_render}
\end{equation}
where $L$ is the rendered pixel color along the view direction $\bc$ of the 3D mesh surface point $\bp$.
$L_i$ is the incident light from the given off-the-shelf environment map, and $\Omega$ is a hemisphere surrounding the surface with the altered surface normal $\bn_\bp$.
Additionally, $f_\theta(\bp, \bc_i, \bc)$ is the bidirectional reflectance distribution function (BRDF) modeled by PBR material modeling, $\bk_d, \bk_{rm}$, and $\bk_n$.
We can split Eq.~\ref{eq:light_render} into diffuse term $L_d$ and the specular term $L_s$ as:
\begin{equation}
\renewcommand{\arraystretch}{2.0}
    \begin{array}{l}
    L(\bp, \bc) = L_d(\bp) + L_s(\bp, \bc), 
    \\
    L_d(\bp) = \bk_d(1-m)\int_{\Omega}L_i(\bp, \bc_i)(\bc_i \cdot \bn_\bp) d\bc_i, 
    \\
    L_s(\bp, \bc) = \bigintss_{\Omega}\dfrac{DFG}{4(\bc \cdot \bn_\bp)(\bc_i \cdot \bn_\bp)}L_i(\bp, \bc_i)(\bc_i \cdot \bn_\bp)d\bc_i,
    \end{array}
\end{equation}
where D, F, and G indicate GGX (\ie, microfacet) distribution, Fresnel term, and statistical light-blocking function, respectively.
Following~\cite{nvdiffrec, chen2023fantasia3d}, the split-sum approximation is used to calculate hemisphere integration.
By merging the pixel colors in the rendered image along the view direction $\bc$, we obtain the rendered image $\bx$, representing the result of the rendering process, denoted as: 
\begin{equation}
   \mathbf{x} =  R(\theta, \bc),
\end{equation}
where $R$ refers to the differentiable renderer~\cite{nvidiffrast} and $\theta$ is the parameters of the MLP network that predict PBR material properties, as depicted in the main paper.
We employ xatlas~\cite{xatlas} for the generation of UV texture maps.
As discussed in~\cite{chen2023fantasia3d}, the integration of sampled 2D textures directly into real graphics engines leads to the emergence of texture seams.

\section{Additional Quantitative Analysis}

In this section, we provide further quantitative comparisons in both cross-domain and in-domain scenarios.
We evaluate cross-domain performance on a subset of the 3D-Front dataset~\cite{fu20213dfront} and in-domain performance on a subset of the Pix3D dataset~\cite{sun2018pix3d}, both of which contain ground-truth 3D meshes.

\subsection{Cross-domain Comparison}
In this section, we provide quantitative comparisons for cross-domain image to shape reconstruction.
We compare the same samples in Table. 2 in the main paper.
For cross-domain comparison, we train all methods, excluding our model $\ours$, on Pix3D~\cite{sun2018pix3d}.
Then, all methods evaluate on 3D-Front~\cite{fu20213dfront}.
$\ours$ shows comparable geometry reconstruction with previous methods, especially in the Chamfer Distance (See ~\Tref{tab:3dfront_fscore}).
It is noteworthy that we do not employ any 3D mesh data in our test-time optimization process. 

\subsection{In-domain Comparison}
We also evaluate our 3D object mesh reconstruction quality at the in-domain scenarios.
Note that our optimization process does not access the ground-truth 3D information, \eg, point clouds, voxels, and meshes, while previous methods~\cite{total3d, zhang2021holisticim3d, liu2022towardsinstpifu, chen2023singlessr} are directly trained with Chamfer Distance with ground-truth meshes as supervision.
Despite this, as shown in~\Tref{tab:pix3d}, $\ours$ shows comparable geometry reconstruction with others.
It is worth noticing that our method also reconstructs image-aligned geometry with realistic textures, whereas others are limited in reconstructing only 3D geometry even trained with 3D shape dataset~\cite{sun2018pix3d}.

\begin{table*}[t]
    \centering
    \small
    \resizebox{0.8\linewidth}{!}{
        \begin{tabular}{c cccccc}
        \toprule
            Metric &  MGN~\cite{total3d} & LIEN~\cite{zhang2021holisticim3d} & InstPIFu~\cite{liu2022towardsinstpifu} & SSR~\cite{chen2023singlessr} & \ours~(Ours) \\
        \midrule
            Chamfer Distance $\downarrow$ & 0.1089 & 0.0975 & 0.0992 & 0.1948 & \textbf{0.0943}\\
            F-Score (\%) $\uparrow$ & 27.32 & 34.29 & 31.65 & 16.51 & 29.96 \\
        \bottomrule
        \end{tabular}
    }
    \vspace{3mm}
    \caption{\textbf{Cross-domain evaluation of feed-forward methods.} 
    We measure the Chamfer Distance and F-Score between the predicted and ground-truth meshes. 
    We conduct the experiment to show the test-time adaptation ability of the unseen test dataset, 3D-Front~\cite{fu20213dfront}.
    Note that although we utilize the icp algorithm, the result of SSR~\cite{chen2023singlessr} could have unexpected errors due to its rotated and translated output geometry results.
    }
    \label{tab:3dfront_fscore}
\end{table*}

\begin{table*}[t]
    \centering
    \small
    \resizebox{0.8\linewidth}{!}{
        \begin{tabular}{c cccccc}
        \toprule
            Metric &  MGN~\cite{total3d} & LIEN~\cite{zhang2021holisticim3d} & InstPIFu~\cite{liu2022towardsinstpifu} & SSR~\cite{chen2023singlessr} & \ours~(Ours) \\
        \midrule
            Chamfer Distance $\downarrow$ & 0.0494 & 0.0319 & 0.0825 & 0.1528 & 0.0612\\
            F-Score (\%) $\uparrow$ & 60.75 & 81.01 & 60.75 & 24.28 & 45.48 \\
        \bottomrule
        \end{tabular}
    }
    \vspace{3mm}
    \caption{\textbf{In-domain evaluation of feed-forward methods.} 
    We measure Chamfer Distance and F-Score between the predicted and ground-truth meshes. 
    We conduct the experiment to show the test-time adaptation ability of Pix3D~\cite{sun2018pix3d}, which is countered when training the Image-to-3D module.
    Note that although we utilize the ICP algorithm, the result of SSR~\cite{chen2023singlessr} could have unexpected errors due to its rotated and translated output geometry results. 
    }
    \label{tab:pix3d}
\end{table*}


\section{Additional Qualitative Analysis}

This section presents additional qualitative analyses due to space constraints in the main paper. 
We provide visual results for 
both in-domain scenarios on the Pix3D dataset~\cite{sun2018pix3d}, as well as the 3D-Front dataset~\cite{fu20213dfront} and real scenes.

\subsection{In-domain Comparison}
\begin{figure*}[p]
  \small
  \centering
    \includegraphics[width=\linewidth]{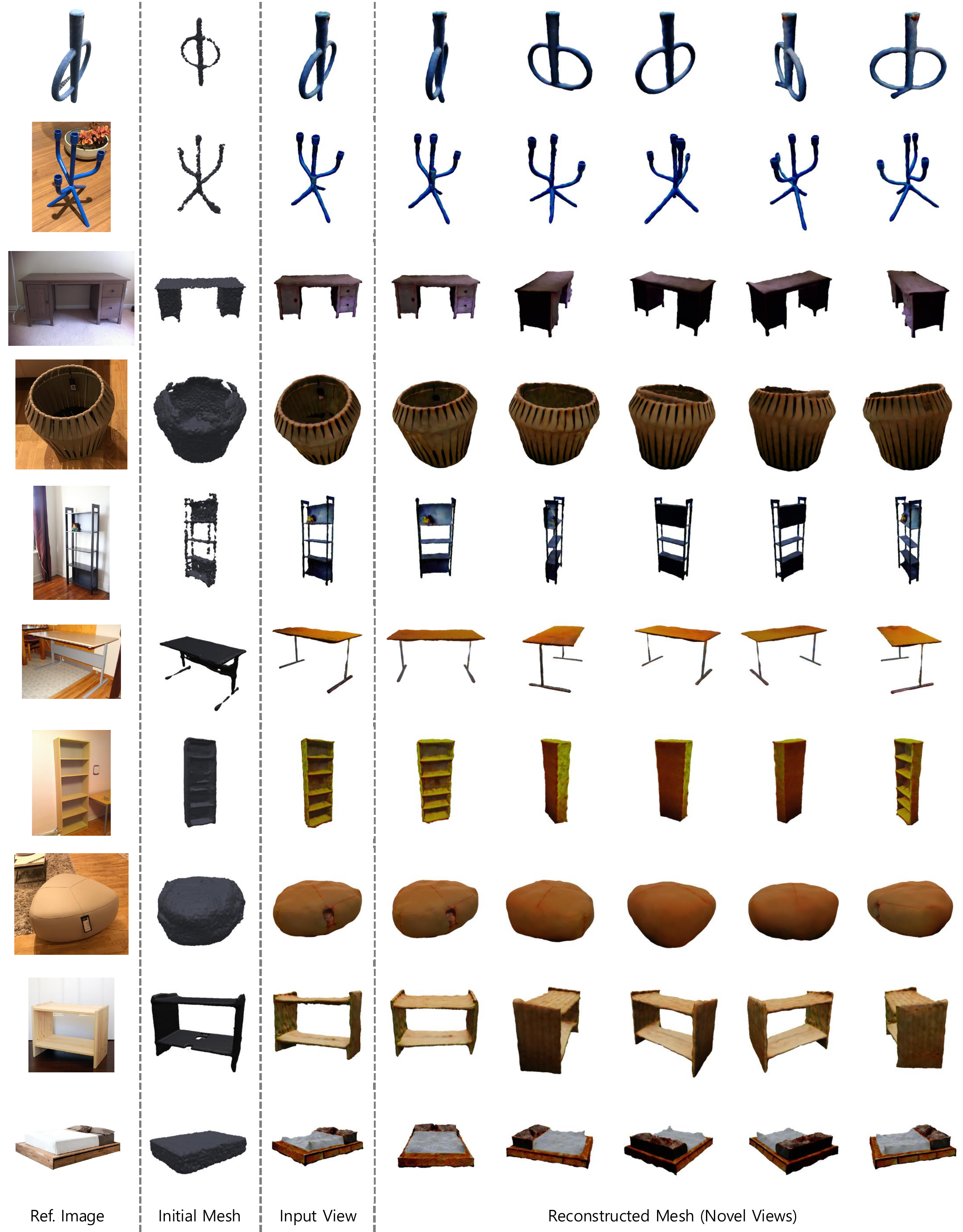}
    \vspace{3mm}
    \caption{\textbf{Additional in-domain experiments about Pix3D~\cite{sun2018pix3d}.}
    We showcase the effectiveness of our test-time adaptation in in-domain scenarios.
    Even in the in-domain settings, the initial mesh prediction is inaccurate with no textures.
    With our test-time adaptation process, we show that fine-grained geometry with realistic textures.
    }
  \label{fig:supp_pix3d1}
\end{figure*}

\begin{figure*}[p]
  \small
  \centering
    \includegraphics[width=\linewidth]{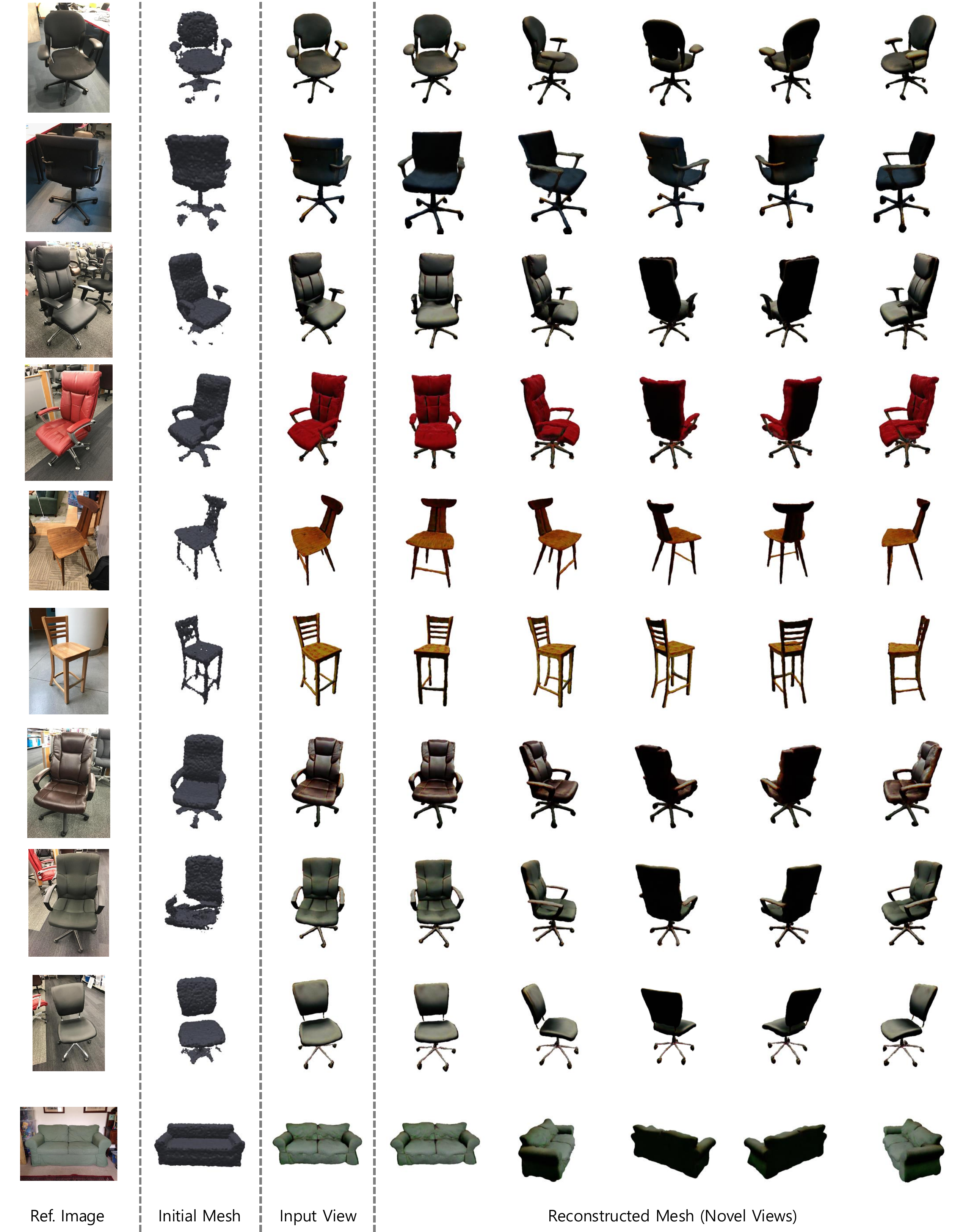}
    \vspace{3mm}
    \caption{\textbf{Additional in-domain experiments about Pix3D~\cite{sun2018pix3d}.}
    We showcase the effectiveness of our test-time adaptation in in-domain scenarios.
    Even in the in-domain settings, the initial mesh prediction is inaccurate with no textures.
    With our test-time adaptation process, we show that fine-grained geometry with realistic textures.
    }
  \label{fig:supp_pix3d2}
\end{figure*}

\begin{figure*}[p]
  \small
  \centering
    \includegraphics[width=\linewidth]{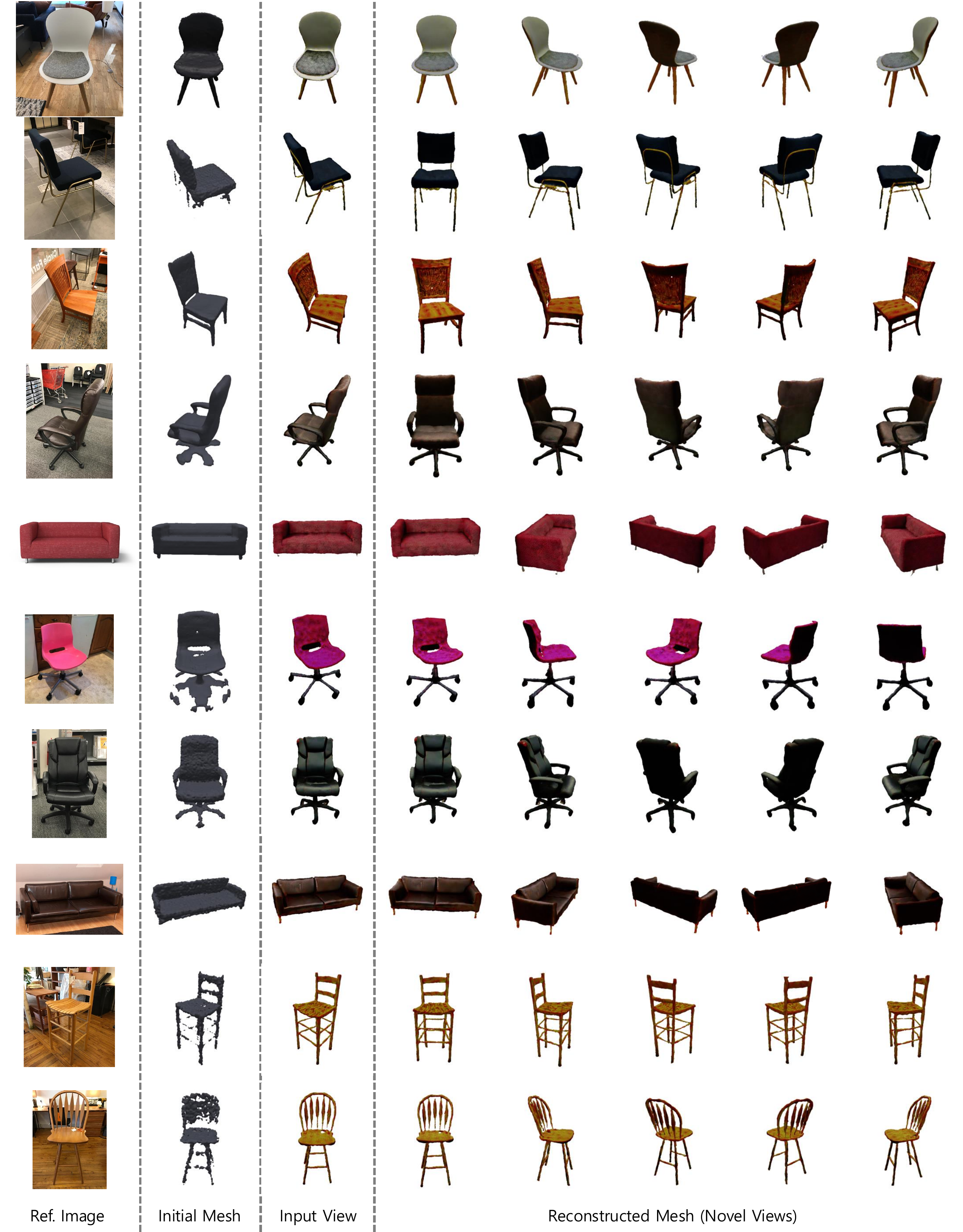}
    \vspace{3mm}
    \caption{\textbf{Additional in-domain experiments about Pix3D~\cite{sun2018pix3d}.}
    We showcase the effectiveness of our test-time adaptation in in-domain scenarios.
    Even in the in-domain settings, the initial mesh prediction is inaccurate with no textures.
    With our test-time adaptation process, we show that fine-grained geometry with realistic textures.
    }
  \label{fig:supp_pix3d3}
\end{figure*}

\begin{figure*}[p]
  \small
  \centering
    \includegraphics[width=\linewidth]{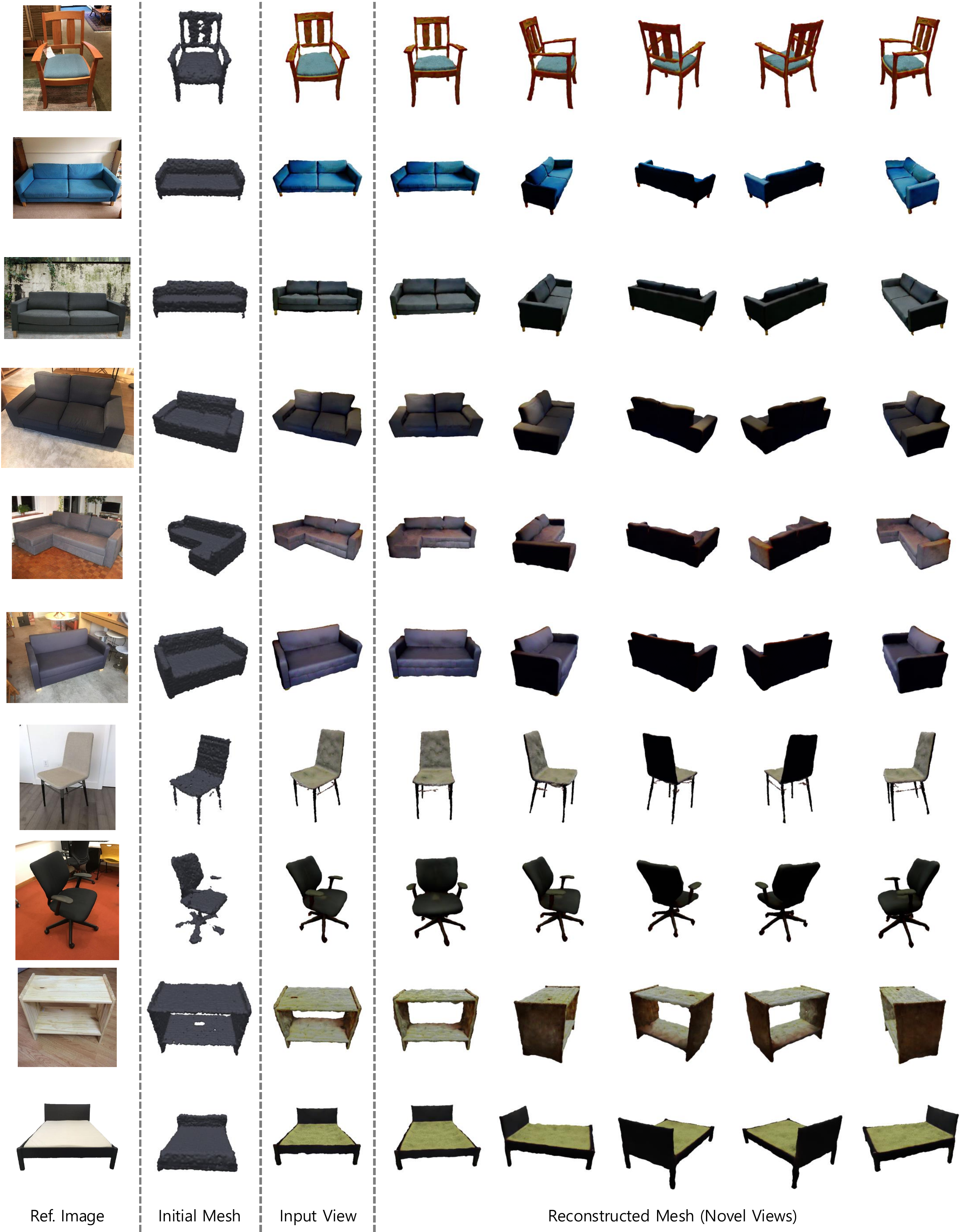}
    \vspace{3mm}
    \caption{\textbf{Additional in-domain experiments about Pix3D~\cite{sun2018pix3d}.}
    We showcase the effectiveness of our test-time adaptation in in-domain scenarios.
    Even in the in-domain settings, the initial mesh prediction is inaccurate with no textures.
    With our test-time adaptation process, we show that fine-grained geometry with realistic textures.
    }
  \label{fig:supp_pix3d4}
\end{figure*}

We assess the performance of our method on the Pix3D dataset, which aligns with our in-domain distribution, resulting in favorable initial mesh predictions as shown in Figs.~\ref{fig:supp_pix3d1},~\ref{fig:supp_pix3d2}, \ref{fig:supp_pix3d3} and~\ref{fig:supp_pix3d4}. 
However, there are instances of erroneous predictions, which our approach effectively rectifies, enhancing the realistic appearance of the reconstruction results.
It is important to note that changes in brightness and contrast may occur due to variations in lighting intensity (\ie, different environment maps).


\subsection{Cross-domain Comparison}
\begin{figure*}[t]
  \small
  \centering
    \includegraphics[width=\linewidth]{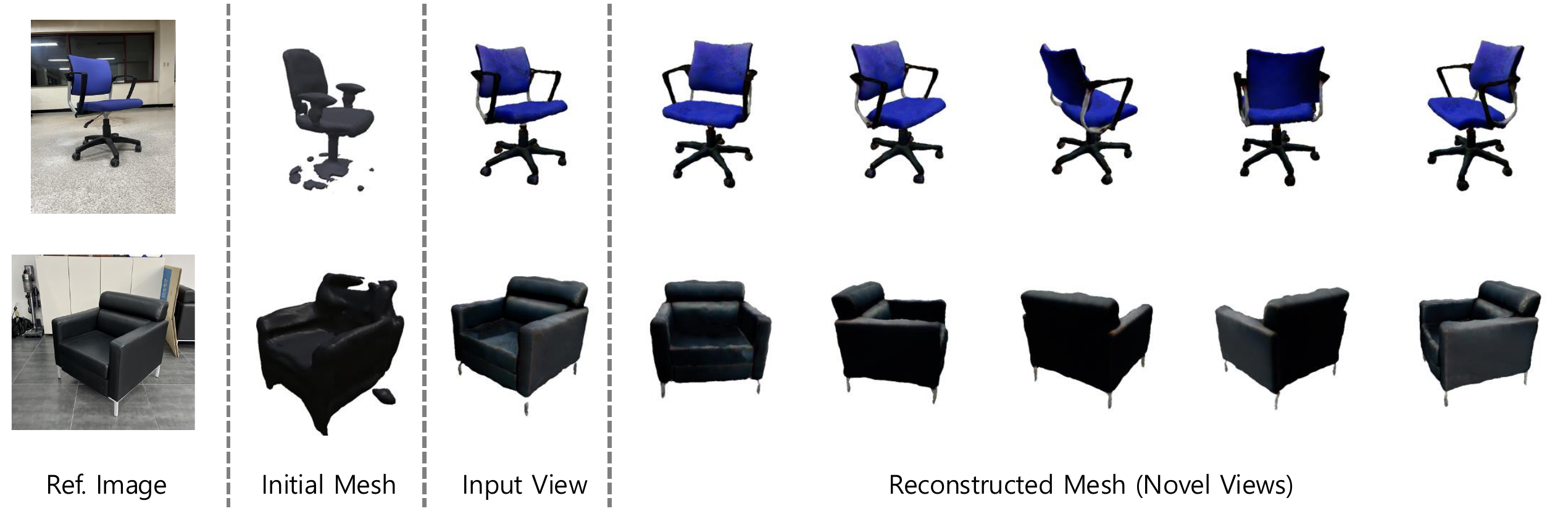}
    \vspace{1mm}
    \caption{\textbf{Additional unseen real-world experiments.}
    We show the additional unseen real-world, \ie, cross-domain experiments with the dataset which we manually acquired.
    }
  \label{fig:supp_real}
\end{figure*}

\begin{figure*}[t]
  \small
  \centering
    \includegraphics[width=\linewidth]{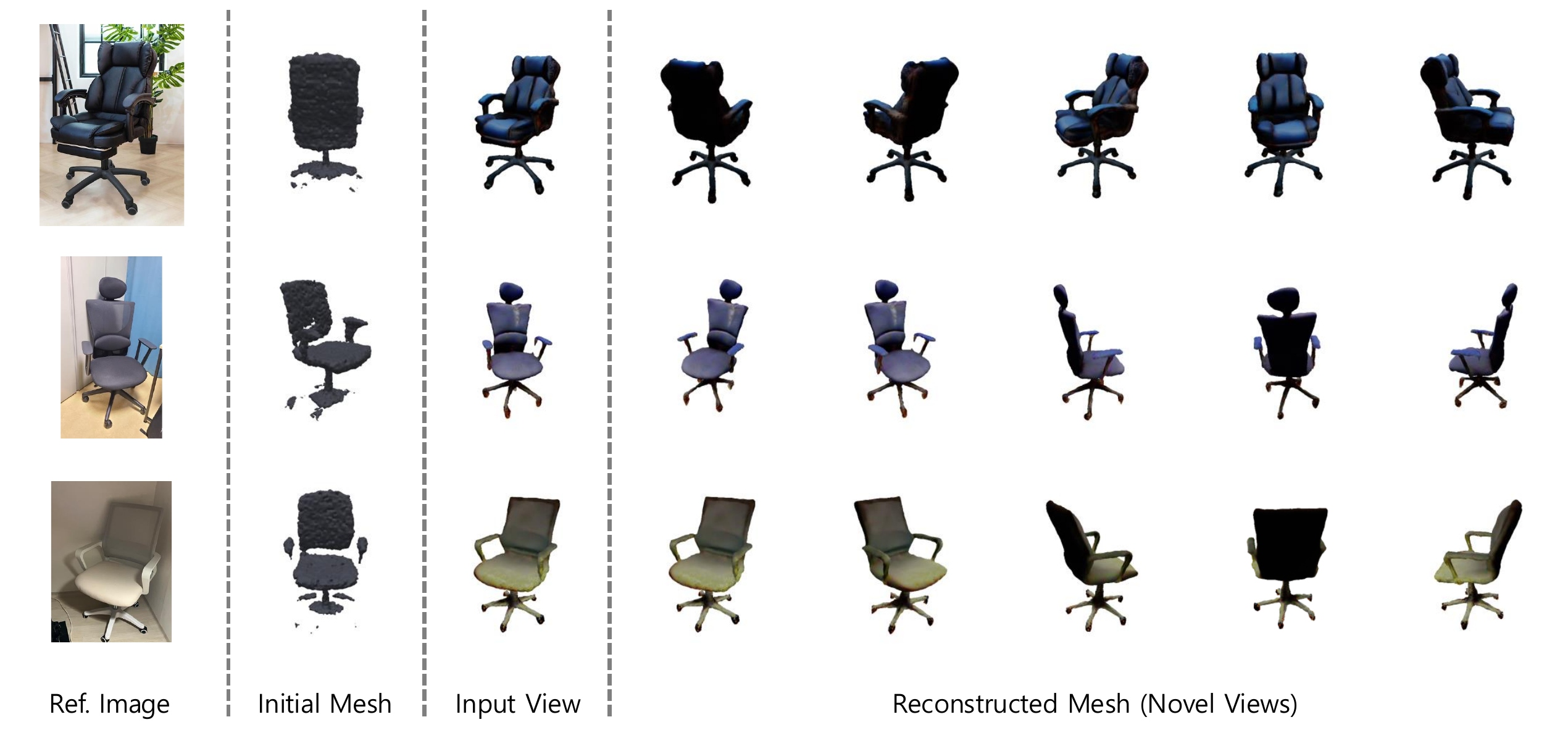}
    \vspace{2mm}
    \caption{\textbf{Additional unseen in-the-wild experiments.} 
    We show the additional in-the-wild, \ie., cross-domain experiments with the dataset we acquired from the web.
    }
  \label{fig:supp_web}
\end{figure*}

We evaluate the performance of an input image from previously unseen distributions through a real scene dataset that we directly acquired and an in-the-wild dataset from the web.
As depicted in Figs.~\ref{fig:supp_real} and~\ref{fig:supp_web}, real-world scenarios represent entirely new domains of images that we have not encountered before.
Consequently, initial mesh predictions struggle to reflect the object shapes within the input image accurately.
However, our test-time adaptation method enables us to obtain fine-grained textured meshes that not only capture the geometry of the input images but also incorporate their textures.

\begin{figure*}[p]
  \small
  \centering
    \includegraphics[width=\linewidth]{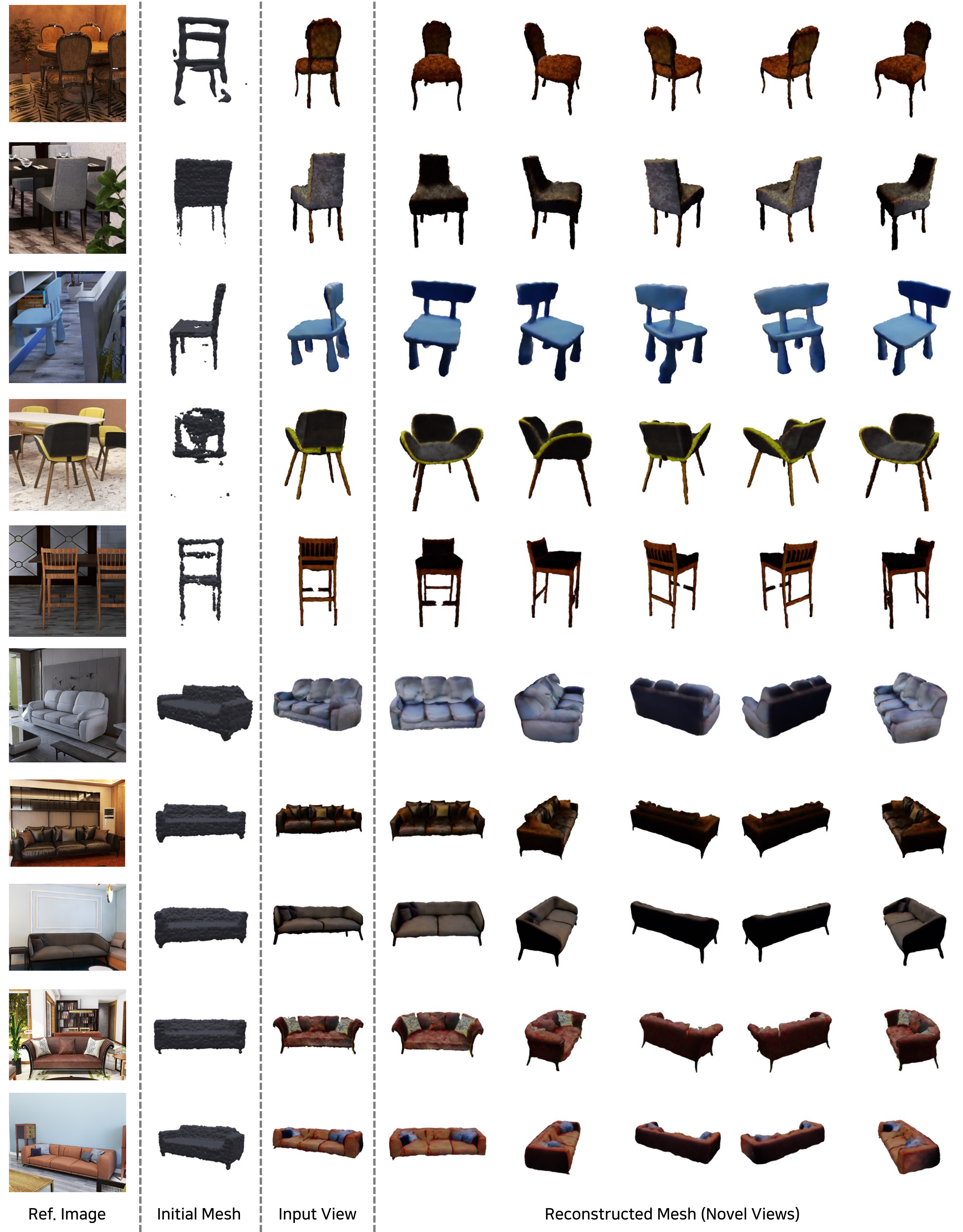}
    \vspace{2mm}
    \caption{\textbf{Additional cross-domain experiments about 3D-Front~\cite{fu20213dfront}.} 
    We showcase the effectiveness of our test-time adaptation in cross-domain scenarios.
    The 3D-Front dataset has not been used in previous feed-forward methods~\cite{total3d, zhang2021holisticim3d}.
    }
  \label{fig:supp_3dfront}
\end{figure*}

We demonstrate the effectiveness of our method on the 3D-Front~\cite{fu20213dfront} dataset, which represents an unseen cross-domain distribution, as illustrated in~\Fref{fig:supp_3dfront}. 
These samples fall outside the training distribution and have not been encountered during training, so initial mesh predictions may not align well with the input image objects. 
However, through our test-time adaptation approach, we can successfully reconstruct object shapes and textures.



\section{In-depth Analysis of Limitation and Discussion}
We conduct in-depth analyses of limitations and discussions that we could not discuss due to the length limitations of the main paper.
Specifically, we present some failure cases of our method and discuss the future direction of improvement.

\begin{figure}[t]
    \centering
    \begin{minipage}{0.48\linewidth}
        \resizebox{1.0\linewidth}{!}{
            \centering
            \includegraphics[width=0.99\linewidth]{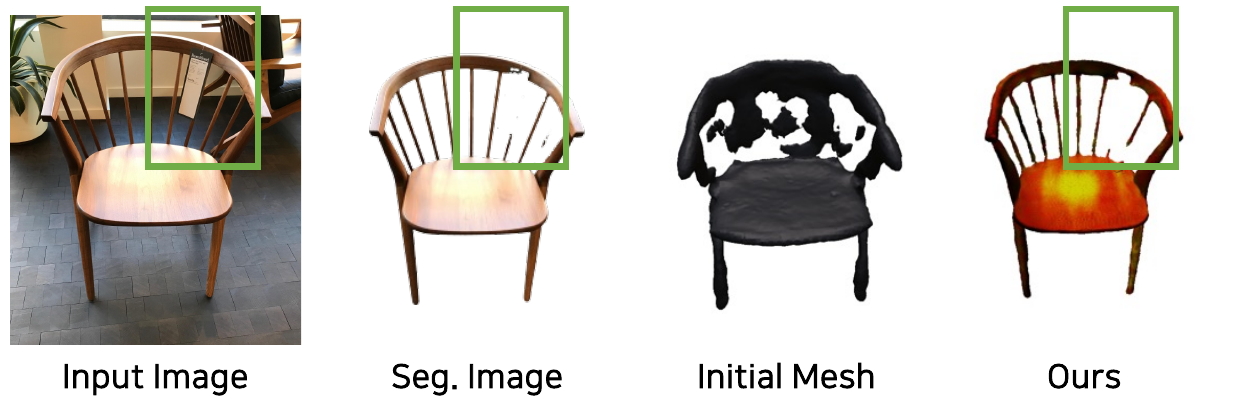}
        }
        \vspace{0.1mm}
        \caption{\textbf{Limitation of model dependencies.} 
        The green square indicates object occlusion in the input image, which disrupts segmentation, leading to the disappearance of the reconstruction mesh.
        }
        \label{fig:supp_segment}
    \end{minipage}
    \hfill
    \begin{minipage}{0.48\linewidth}
        \resizebox{1.0\linewidth}{!}{
            \centering
            \includegraphics[width=0.99\linewidth]{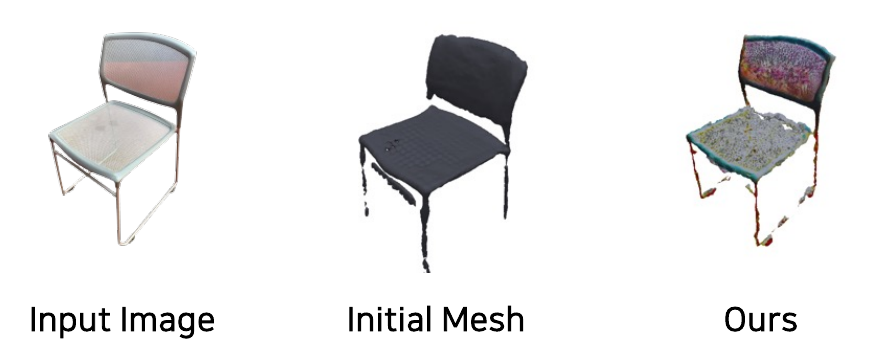}
        }
        \vspace{0.1mm}
        \caption{\textbf{Limitation of transparency or reflection surface.} 
        The surface texture of the input image is transparent and features special material properties of the mesh material.
        As a result, both the output geometry and texture are degraded.
        }
        \label{fig:supp_pbr}
    \end{minipage}
\end{figure}

\subsection{Failure Cases}\label{sec:fail}
\paragraph{Model dependencies}
Our model utilizes the initial mesh and viewpoint predictions from the Image-to-3D module as the starting point for single-view image to 3D textured mesh reconstruction. 
It implies that our single-view to 3D capabilities are constrained by the capacity of the Image-to-3D module (\eg, it only functions for categories where viewpoint prediction is feasible). 
Furthermore, we require images segmented to include only the object of interest to utilize the multi-view diffusion model. 
Therefore, the quality of segmentation directly impacts the quality of 3D reconstruction as shown in~\Fref{fig:supp_segment}.

\paragraph{Transparency or reflection surface}
Reconstructing 3D objects from single-view images has been a long-standing challenge. 
In addition, estimating PBR (Physically-Based Rendering) materials from single-view images presents an ill-posed problem, as there is inherent ambiguity between the diffuse component and lighting.
In particular, the models currently in use assume microfacet surfaces~\cite{walter2007microfacet}. 
Therefore, for instances with special material properties involving transparency or reflection, the texture optimization tends to degrade, resulting in sub-optimal geometry updates as showin in~\Fref{fig:supp_pbr}.

\subsection{Discussion}
\begin{wrapfigure}{r}{0.5\linewidth}
    \small
    \vspace{-6mm}
    \centering
        \includegraphics[width=0.99\linewidth]{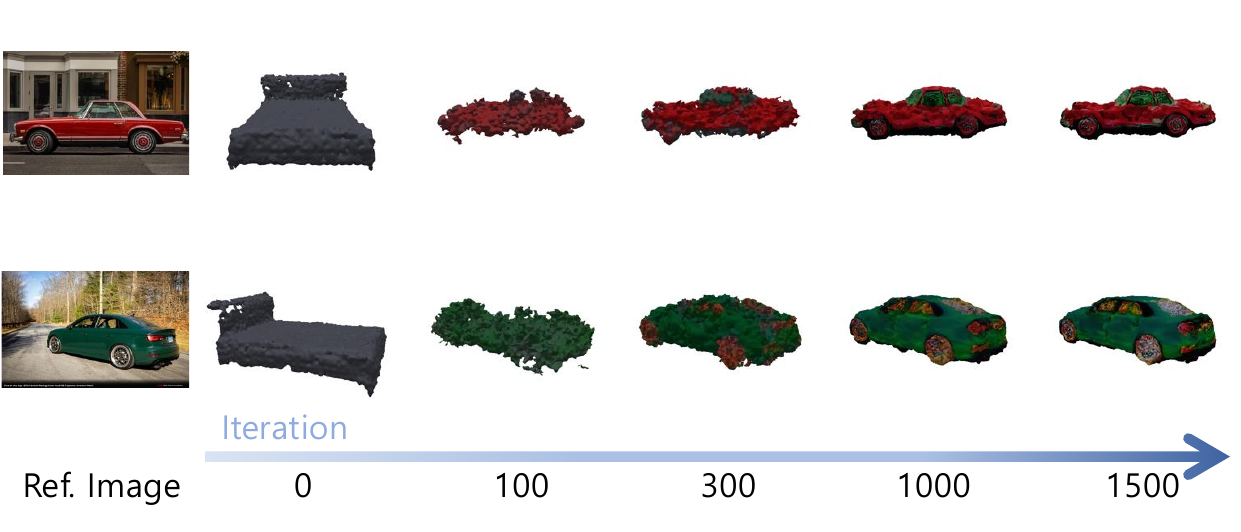}
        \vspace{1mm}
        \caption{
        \textbf{Possibility of category extension.}
        }
    \label{fig:supp_general_class}
    \vspace{-4mm}
\end{wrapfigure}
We believe that expanding the Image-to-3D module into a more robust one capable of handling a larger class vocabulary could overcome model dependency issues despite the dependencies on the model in use.
Because our test-time adaptation stage has the capability to category generalization as shown in~\Fref{fig:supp_general_class}.
Additionally, addressing the degradation in reconstruction quality due to special reflection surfaces might be achievable through further exploration and application of complex material modeling and rendering equations in the future.
Our research is practical in that it introduces a pipeline capable of operating in previously unseen out-of-distribution scenarios, especially in real-scene scenarios, which were not extensively considered in prior studies and can work for various viewpoint conditions in real images, which is different from existing generative prior methods~\cite{melas2023realfusion, liu2023zero123, tang2023makeit3d}.
We believe that our work can serve as a stepping stone for the advancement of single-view to 3D reconstruction methods that operate effectively in real scenarios.

\end{document}